\documentclass[runningheads]{llncs}
\usepackage{graphicx}
\usepackage{comment}
\usepackage{amsmath,amssymb} % define this before the line numbering.
\usepackage{color}

\usepackage[width=122mm,left=12mm,paperwidth=146mm,height=193mm,top=12mm,paperheight=217mm]{geometry}

\usepackage{hyperref}       % hyperlinks
\usepackage{url}            % simple URL typesetting
\usepackage{booktabs}       % professional-quality tables
\usepackage{amsfonts}       % blackboard math symbols
\DeclareMathOperator*{\argmax}{arg\,max}

\usepackage{bm}

\usepackage{algorithm}
\usepackage{algorithmic}

\usepackage{caption}
\usepackage{subfigure}
\usepackage{multirow}
\usepackage{makecell}
\newcommand\mgape[1]{\gape{$\vcenter{\hbox{#1}}$}}
\newcommand{\beq}{\begin{equation}\begin{aligned}}
\newcommand{\eeq}{\end{aligned}\end{equation}}
\newcommand{\beqn}{\begin{equation*}\begin{aligned}}
\newcommand{\eeqn}{\end{aligned}\end{equation*}}

\makeatletter
\newcommand{\thickhline}{%
	\noalign {\ifnum 0=`}\fi \hrule height 1pt
	\futurelet \reserved@a \@xhline
}
\newcolumntype{"}{@{\hskip\tabcolsep\vrule width 0.8pt\hskip\tabcolsep}}
\newcommand{\thickcline}[1]{%
	\@thickcline #1\@nil%
}

\def\@thickcline#1-#2\@nil{%
	\omit
	\@multicnt#1%
	\advance\@multispan\m@ne
	\ifnum\@multicnt=\@ne\@firstofone{&\omit}\fi
	\@multicnt#2%
	\advance\@multicnt-#1%
	\advance\@multispan\@ne
	\leaders\hrule\@height1pt\hfill
	\cr
	\noalign{\vskip-1pt}%
}
\makeatother

\usepackage{setspace}

\begin{document}
% \linenumbers
\pagestyle{headings}
\mainmatter
\def\ECCVSubNumber{}  % Insert your submission number here

\title{Training few-shot classification via the perspective of minibatch and pretraining} % Replace with your title

%\begin{comment}
\titlerunning{Training few-shot classification} 
%\authorrunning{ECCV-20 submission ID \ECCVSubNumber} 
\author{Meiyu Huang \and Xueshuang Xiang \and Yao Xu}
\institute{Qian Xuesen Laboratory of Space Technology \\ China Academy of Space Technology}
%\end{comment}
%******************

\begin{comment}
\titlerunning{Abbreviated paper title}

\author{First Author\inst{1}\orcidID{0000-1111-2222-3333} \and
Second Author\inst{2,3}\orcidID{1111-2222-3333-4444} \and
Third Author\inst{3}\orcidID{2222--3333-4444-5555}}

%
\authorrunning{F. Author et al.}
\institute{Princeton University, Princeton NJ 08544, USA \and
Springer Heidelberg, Tiergartenstr. 17, 69121 Heidelberg, Germany
\email{lncs@springer.com}\\
\url{http://www.springer.com/gp/computer-science/lncs} \and
ABC Institute, Rupert-Karls-University Heidelberg, Heidelberg, Germany\\
\email{\{abc,lncs\}@uni-heidelberg.de}}
\end{comment}
%******************
\maketitle

\begin{abstract}
Few-shot classification is a challenging task which aims to formulate the ability of humans to learn concepts from limited prior data and has drawn considerable attention in machine learning. Recent progress in few-shot classification has featured meta-learning, in which a parameterized model for a learning algorithm is defined and trained to learn the ability of handling classification tasks on extremely large or infinite episodes representing different classification task, each with a small labeled support set and its corresponding query set. In this work, we advance this few-shot classification paradigm by formulating it as a supervised classification learning problem. We further propose multi-episode and cross-way training techniques, 
which respectively correspond to the minibatch and pretraining in classification problems. 
Experimental results on a state-of-the-art few-shot classification method (prototypical networks) demonstrate that both the proposed training strategies can highly accelerate the training process without accuracy loss for varying few-shot classification problems on Omniglot and \emph{mini}ImageNet.
\keywords{Few-shot classification, multi-episode traing, cross-way training}
\end{abstract}

\section{Introduction}

Few-shot classification problems~\cite{Lake2011One,Li2006One} have drawn much attention, since they can formulate the ability of humans to learn concepts from limited prior data. Formally, from the view of classification, few-shot classification is a challenging task that aims to learn information about object categories from one or only a few labeled samples. To overcome the drawbacks of adopting the deep learning techniques used in large-scale classification problems to address few-shot classification, a well-established paradigm is to train few-shot classification in an auxiliary meta-learning~\cite{schmidhuber1987evolutionary,naik1992meta} or learning-to-learn~\cite{thrun2012learning,Hochreiter2001Learning} phase, where 
\iffalse
transfer learning is performed from a pool of various classification tasks generated from large quantities of available labeled data, to new classification tasks from classes unseen at training time. And 
\fi
transferrable knowledge is learned in the form
of good initial conditions using optimization based methods~\cite{FinnAL2017Model,ravi2017optimization,Nichol2018on,rusu2018meta} or embeddings using memory based methods~\cite{santoro2016meta,munkhdalai2017meta,Mishra2017Meta} and metric based methods~\cite{koch2015siamese,vinyals2016matching,shyam2017attentive,snell2017prototypical,sung2018learning,DBLP:journals/corr/abs-1803-00676}. The target few-shot classification problem
is then learned by fine-tuning~\cite{FinnAL2017Model,Nichol2018on,rusu2018meta} with the learned optimization
strategy~\cite{ravi2017optimization} or computed in a feed-forward pass
~\cite{santoro2016meta,munkhdalai2017meta,koch2015siamese,vinyals2016matching,shyam2017attentive,snell2017prototypical,sung2018learning,DBLP:journals/corr/abs-1803-00676} without updating network weights.

These various meta-learning formulations have led to significant progress recently in few-shot classification. And the best performing methods prescribed by meta-learning use the training framework based on episodes~\cite{vinyals2016matching}, each with a small labeled support set and its corresponding query set generated from large quantities of available labeled data, to mimic the few-shot setting in the test environment for improving generalization. In this episodic training framework, few-shot classification can be viewed as learning the ability of \emph{classifying an unlabeled query example given a small labeled support set} (a classification task) by training on a large number of classification tasks ($C^5_{1200}$ kinds of tasks if we consider the $5$-way few-shot classification problem on Omniglot~\cite{Lake2011One} with 1200 training classes). From this point of view, the training data is not limited but extremely massive. 

\begin{figure*}[htb]
	\centering
	\includegraphics[width=0.9\textwidth]{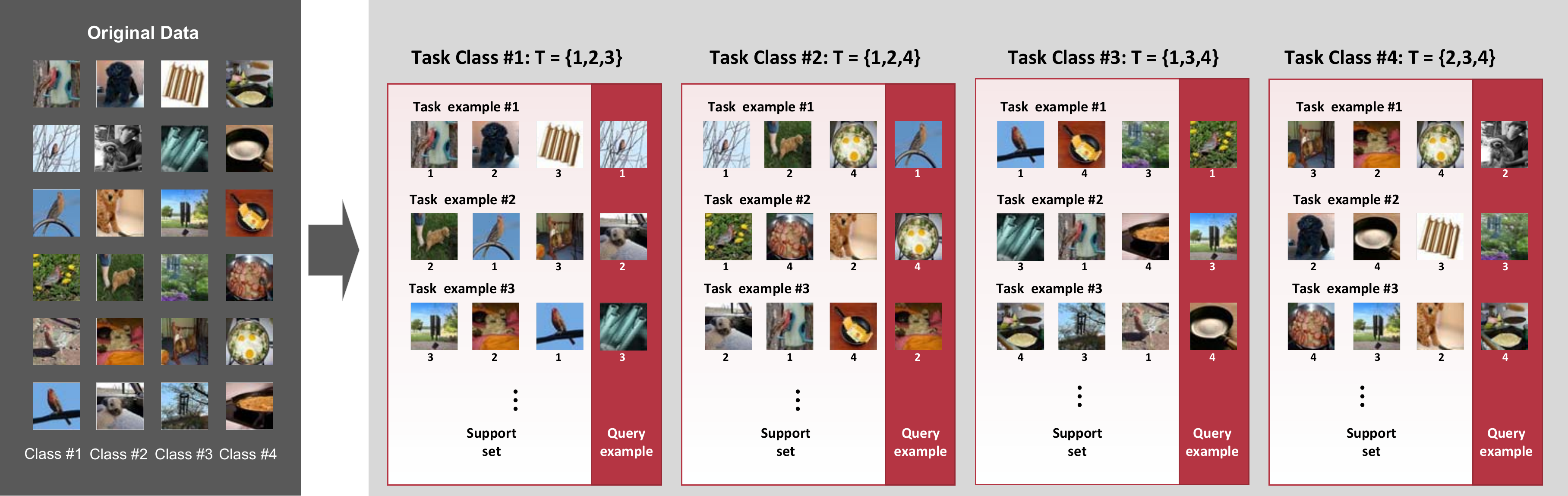}
	%%\vspace{-0.1in}
	\caption{An illustration of task classes and task examples for a $3$-way $1$-shot classification problem on a dataset with $4$ classes, where each class has $6$ examples. By the definition of task class and task example, we have $C_4^3=4$ task classes and $(C_6^1)^3C_3^1(6-1)=3240$ task examples in each task class. From the view of supervised learning, we may analogize the task class and task examples of few-shot classification problems (the right part of the figure) to the class and examples of regular classification problems (the left part of the figure), respectively. Thus, the classification problem aims to learn the ability of classification, while the few-shot classification problem aims to learn the ability of handling classification tasks.}
	%\vspace{-0.1in}
	\label{fig:def}
\end{figure*}

\begin{figure*}[htb]
	\centering
	\includegraphics[width=0.75\textwidth]{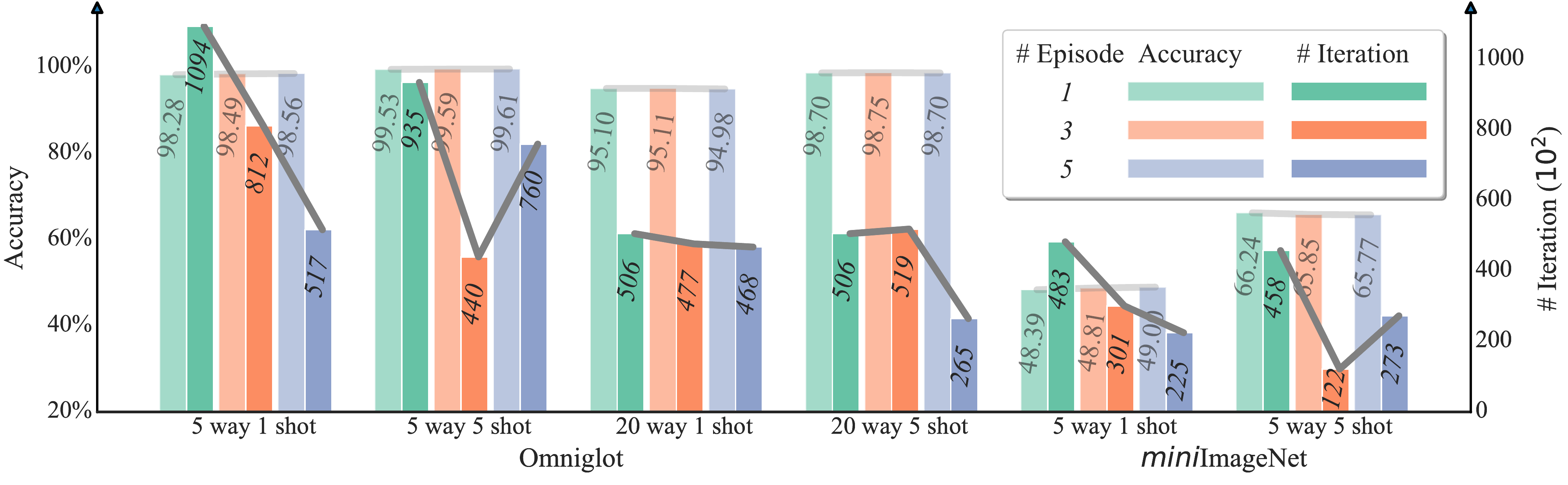} 
	\includegraphics[width=0.75\textwidth]{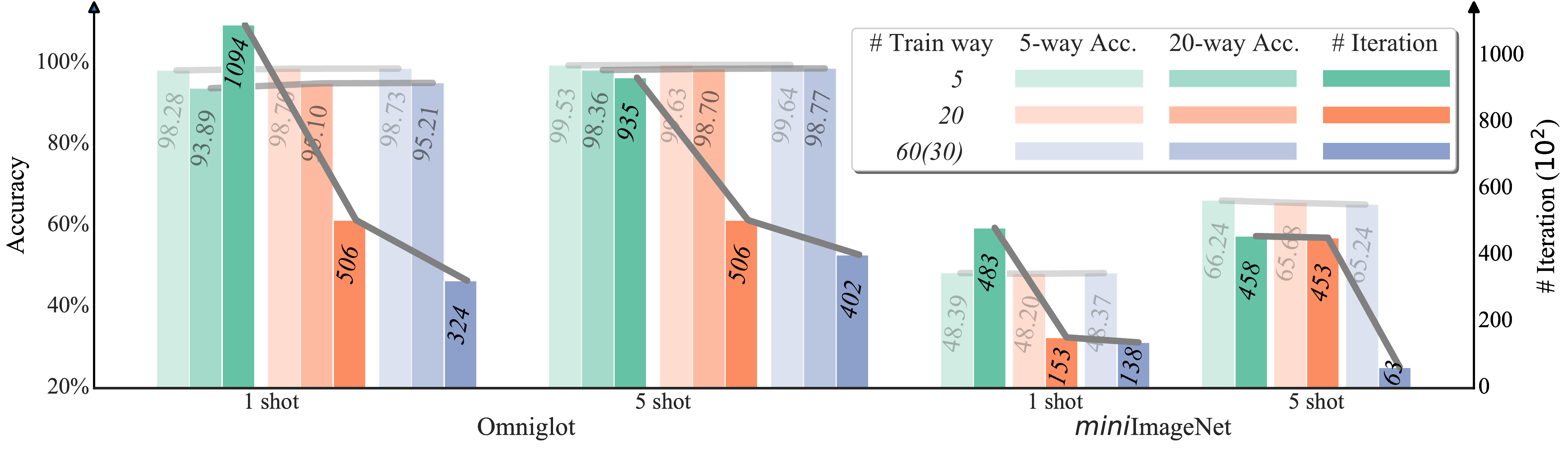} 
	\caption{The convergence accuracy-iterations results of \textbf{(Top)} multi-episode training and \textbf{(Bottom)} cross-way training on Prototypical Networks~\cite{snell2017prototypical} (with Euclidean distance) with the optimum learning rate decaying policy for Omniglot and \emph{mini}ImageNet.}
	\label{fig:multi-episode-acc-iter}
\end{figure*}

This paper advances this view by explaining it in a way that is more explicit or formal, and further transfers the learning approach of classification problem into few-shot learning area. 
%To be specific, we first formulate classification tasks as examples and then parallel few-shot classification to regular supervised classification (as shown in Figure~\ref{fig:def}). On top of the formalism of the parallel, we further propose the multi-episode and cross-way training strategies for few-shot classification, which respectively correspond to the minibatch and pretraining strategies in regular classification problems. We then implement the two proposed training strategies on Prototypical Networks~\cite{snell2017prototypical}, a state-of-the-art metric based few-shot classification method. Experimental results demonstrate that both training strategies help accelerate the training process without accuracy loss for the few-shot classification problems of Omniglot and \emph{mini}ImageNet. 
The main contribution of this paper can be summarized as follows:
\begin{itemize}
	\item Inspired by the episodic training paradigm~\cite{vinyals2016matching}, we introduce a formal parallel between regular supervised classification and few-shot classification (as shown in Figure~\ref{fig:def}), which motivates us to improve the training strategies of few-shot classification. 
	%propose two variants of training strategies for few-shot classification. 
	\item Compared to minibatch (several examples in each class in each iteration) training in regular classification problems, we first suggest using multi-episode (several task examples in multiple task class in each iteration) training in metric based few-shot classification methods. 
	%Experimental results on Prototypical Networks~\cite{snell2017prototypical} demonstrate that multi-episode training helps speed up convergence on the target few-shot classification task compared with one-episode (several task examples in one task class in each iteration) training~\cite{vinyals2016matching} that requires a slower learning rate decaying policy and more iterations to sufficiently converge.
	%, because multi-episode training increases the mini-batch size and the degree of parallel computing
	%Moreover, multi-episode training can achieve better performance on most target tasks than one-episode training without architectural changes. 
	%The reason may be that multi-episode training can well relieve the problem of imbalanced task class sampling in minibatch selection in one-episode training. These results suggest researchers would be free to use multi-episode training to reduce computation time in few-shot classification. 
	\item Compared to the pretraining (training the model on a similar dataset with large-scale data) technique in regular classification problems, such as ImageNet pretraining~\cite{Russakovsky2015ImageNet}, we first suggest using cross-way training (pretraining the model on a task with a higher way) in few-shot classification problems. 
	%Experimental results on Prototypical Networks~\cite{snell2017prototypical} demonstrate that pretraining on a few-shot problem with a higher way converges faster than training on the target problem of a lower way and helps improve the testing accuracy on the target problem. 
	%This result may occur because there are more data in each episode when training with a higher way, and pretraining with a higher way can generate a more ``universal'' feature representation as ImageNet pretraining. 
\end{itemize}
Experimental results on Prototypical Networks~\cite{snell2017prototypical} demonstrate that the proposed multi-episode and cross-way training strategies help speed up 
the convergence or even improve the testing accuracy on the target few-shot classification problems, see Figure~\ref{fig:multi-episode-acc-iter} for details. 

\section{Related work}
Recent works on few-shot classification can be mainly categorized into three classes, which are optimization-based methods~\cite{FinnAL2017Model,ravi2017optimization,Nichol2018on,rusu2018meta}, memory-based methods~\cite{santoro2016meta,munkhdalai2017meta,Mishra2017Meta} and metric-based methods~\cite{koch2015siamese,vinyals2016matching,shyam2017attentive,snell2017prototypical,sung2018learning}.

The approach within which our work falls is that of metric-based methods. Previous work in metric-learning for few-shot-classification includes Deep Siamese Networks~\cite{koch2015siamese}, Matching Networks~\cite{vinyals2016matching}, Relation Networks~\cite{sung2018learning}, and Prototypical Networks~\cite{snell2017prototypical}, which is the model we implement the two proposed training strategies in our work. The main idea here is to learn an embedding function that embeds examples belonging to the same class close together while keeping embeddings from different classes far apart. Distances between embeddings of examples from the support set and query set are then used as a notion of similarity to do classification. Lastly, closely related to our work, Matching Networks notably introduced the training framework based on episodes representing different classification tasks, which inspires us to formulate classification tasks as examples and parallel few-shot classification with regular classification. On top of the formal parallel, we propose the multi-episode training strategy, which can reduce training iterations before convergence without accuracy loss compared with the one-episode training strategy used in Matching Networks and the follow-up metric-based methods~\cite{snell2017prototypical,sung2018learning}. 

\iffalse
The other two kinds of few-shot classification approaches, namely optimization-based methods and memory-based methods include learning how to use the support set to update a learner model so as to generalize to the query set. Specifically, optimization-based methods has involvoded learning either the weight initialization and/or update step that is used by a learner neural network~\cite{FinnAL2017Model,ravi2017optimization,Nichol2018on,rusu2018meta}. Memory-based methods are to train generic neural architecture such as a memory-augmented recurrent network~\cite{santoro2016meta} or a temporal convolutional network~\cite{Mishra2017Meta} to sequentially process the support set and perform accurate predictions of the labels of the query set examples. These other methods are also competitive for few-shot learning, but we chose to extend Prototypical Networks in this work for its simplicity and efficiency.
\fi

The other two kinds of few-shot classification approaches, namely optimization-based methods and memory-based methods include learning how to use the support set to update a learner model so as to generalize to the query set.
Specificaly, optimization-based approaches aim to optimize the model for fast adaptability, allowing it to adapt to new tasks with only a few examples. MAML~\cite{FinnAL2017Model} and REPTILE~\cite{Nichol2018on} reach this target by learning a set of parameters of a given neural network for sensitivity on a given task distribution, so that it can be efficiently fine-tuned for a sparse data problem within a few gradient-descent update steps. To bypass the limitation of MAML when operating on high-dimensional parameter spaces in extreme low-data regimes, Rusu et al.~\cite{rusu2018meta} proposed LEO, which first learns a data-dependent latent generative representation of model parameters and then performs gradient-based meta-learning in this low dimensional latent space. Instead of only training a good initial condition, Ravi et al.~\cite{ravi2017optimization} introduced an LSTM~\cite{hochreiter1997long}-based optimizer that is trained to be specifically effective for fine-tuning. Memory-based methods~\cite{santoro2016meta,munkhdalai2017meta,Mishra2017Meta}, on the other hand, exploit memory neural network architectures such as the RNN~\cite{mikolov2010recurrent} to store and retrieve knowledge in their memories to fulfill the given tasks. For instance, the work of memory-augmented neural networks (MANNs)~\cite{santoro2016meta} learns quickly from data presented sequentially with an LSTM architecture. These other methods are also competitive for few-shot learning, but we chose to extend Prototypical Networks in this work for its simplicity and efficiency. 

\section{Methodology}
\label{sec:methodology}
This section will first introduce the background including the episodic training paradigm and 
Prototypical Networks~\cite{snell2017prototypical}. 
Then we will give the formal parallel of the regular classification problem and the few-shot classification problem from the view of supervised learning. 
Motivated by the formal parallel, we finally propose the corresponding multi-episode and cross-way training strategies for few-shot classification problems via the perspective of minibatch training and pretraining.
%Then, we will compare few-shot classification to regular classification via the perspective of minibatch training and pretraining. At each training strategy, we will start from the regular classification problem and then introduce the corresponding multi-episode and cross-way training for few-shot classification problems. 
Table~\ref{tab:comparison} shows a detailed comparison between regular classification and few-shot classification, where the loss function proposed in Prototypical Networks~\cite{snell2017prototypical} is used for few-shot classification for instance.
\begin{table*}[!htb]
	\centering
	\caption{Comparisons between regular classification and few-shot classification from the view of supervised learning. We use the loss function proposed in Prototypical Networks~\cite{snell2017prototypical} for few-shot classification for instance.}\label{tab:comparison}
	\begin{tiny}
		\scalebox{0.95}[0.95]{
			\begin{tabular}{p{0.08\textwidth}<{\centering}p{0.42\textwidth}<{\centering}p{0.45\textwidth}<{\centering}}
				\specialrule{1pt}{0pt}{2pt}
				&\textbf{Classification} 	&\textbf{$K$-way $S$-shot learning}\\
				\specialrule{0.8pt}{2pt}{2pt}
				\textbf{Dataset}& $\scriptstyle \mathcal{D} =\{ (\mathbf{x}_i, y_i)\}_{i=1}^{LH}=\cup_{j=1}^L\mathcal{D}_j$ &	$\mathcal{D}^f =\{(\bm{\tau}_i, y_i)\}_{i=1}^{|\mathcal{T}||\mathcal{G}({V})|} =\cup_{V \in \mathcal{T}}\, \mathcal{G}({V})$,\\
				& & 	where $\scriptstyle \bm{\tau}_i=\{\mathcal{S}_V=\cup_{k=1}^K\mathcal{S}_{V_k},\mathbf{x}_{i}\}$\\
				& & 
				$\scriptstyle |\mathcal{T}| = C_{L}^K$,\quad $\scriptstyle |\mathcal{G}({V})|=(C_{H}^S)^KC_K^1(H-S)$\\
				\specialrule{0.5pt}{3pt}{3pt}
				\textbf{Minibatch Training} &	$\bm{\theta}_{t+1}=\bm{\theta}_t-{\alpha}\sum_{(\mathbf{x}_i, y_i) \in \mathcal{B}_t} \frac{\partial l(f_{\bm{\theta}};\mathbf{x}_i,y_i)}{\partial \bm{\theta}}\, |_{\bm{\theta}=\bm{\theta}_t},$& $\bm{\theta}_{t+1}=\bm{\theta}_t-{\alpha}\sum_{(\bm{\tau}_i, y_i) \in \mathcal{B}_t} \frac{\partial l(f_{\bm{\theta}};\bm{\tau}_i,y_i)}{\partial \bm{\theta}}\, |_{\bm{\theta}=\bm{\theta}_t},$\\
				& $l(f_{\bm{\theta}};\mathbf{x}_i,y_i)=-\log p(y=y_i|\mathbf{x}_i),$ &  $l(f_{\bm{\theta}};\bm{\tau}_i,y_i)=-\log p(y=y_i|\bm{\tau}_i),$\\
				&$p(y=j|\mathbf{x}_i)=\frac{\exp{(-f_{\bm{\theta}}(\mathbf{x}_i)_j)}}{\sum_{j'=1}^L \exp{(-f_{\bm{\theta}}(\mathbf{x}_i)_{j'})}}$ &$p(y=V_k|\bm{\tau}_i)=\frac{\exp{(-d(f_{\bm{\theta}}(\mathbf{x}_i),\mathbf{c}_{V_k}))}}{\sum_{k'=1}^K \exp{(-d(f_{\bm{\theta}}(\mathbf{x}_i),\mathbf{c}_{V_{k'}}))}},$\\
				& &
				$\mathbf{c}_{V_k}=\frac{1}{|\mathcal{S}_{V_k}|}\sum_{(\mathbf{x}_{i'},y_{i'})\in \mathcal{S}_{V_k}} f_{\bm{\theta}}(\mathbf{x}_{i'})$\\
				& Randomly select several examples in $\mathcal{D}$ uniformly as $\mathcal{B}_t$. 
				%Suppose $|\mathcal{B}_t|=100$ and $M=10$, then in the sense of probability, 
				%$\mathcal{B}_t$ will have about $10$ examples from each $\mathcal{D}_j$. 
				& Sampling several task classes in $\mathcal{T}$ and then sampling several task examples of the sampled task classes as a minibatch, namely, $\mathcal{B}_t:= \cup_{e=1}^E \mathcal{B}_t^e$, where $\mathcal{B}_t^e \subset \mathcal{G}(V^e)$ is an episode~\cite{vinyals2016matching}.
				\\
				\specialrule{0.5pt}{3pt}{3pt}
				\textbf{Pre-training} &
				$\bm{\theta}_0={\rm argmin}_{\bm{\theta}} \, \sum_{(\mathbf{x}_i, y_i) \in \mathcal{D}_{\rm pre}} l(f_{\bm{\theta}};\mathbf{x}_i, {y}_i),$  &	$\bm{\theta}_0={\rm argmin}_{\bm{\theta}} \, \sum_{(\bm{\tau}_i, y_i) \in \mathcal{D}_{\rm pre}} l(f_{\bm{\theta}};\bm{\tau}_i, {y}_i),$\\
				&  $|\mathcal{D}| < |\mathcal{D}_{\rm pre}=ImageNet|$ &$|\mathcal{D}^f|
				%=C_{L}^K (C_{H}^S)^KC_K^1(H-S) 
				< |\mathcal{D}_{\rm pre}=\mathcal{D}^{\hat K{\rm -way} S {\rm -shot}}|$
				%=C_{L}^{\hat K} (C_{H}^S)^{\hat K}C_{\hat K}^1(H-S)$
				, $ \hat K > K$\\
				\specialrule{1pt}{2pt}{0pt}	
			\end{tabular}
		}
	\end{tiny}
\end{table*}

\subsection{Background}
%We start by defining precisely the current episodic training paradigm for few-shot learning and the Prototypical Networks~\cite{snell2017prototypical} approach to this problem.
{\bf Episodic training paradigm}. 
Suppose we consider a few-shot classification problem on a large labeled dataset with $L$ classes and each class of $H$ examples denoted by $\mathcal{D} =\{ (\mathbf{x}_i, y_i)\}_{i=1}^{LH}=\cup_{j=1}^L\mathcal{D}_j$, where $\mathbf{x}_i \in \mathbb{R}^D$ is an input vector of dimension $D$, $y_i \in\{1,2,\cdots,L\}$ is a class label, $\mathcal{D}_j$ denotes the subset of $\mathcal{D}$ containing all elements $(\mathbf{x}_i, y_i)$ 
such that $y_i = j$.
The ultimate goal of the episodic paradigm is to produce classifiers for a disjoint set $\mathcal{D}_{test}$ %with $S$ labelled examples for each of $K$ unique classes.
of new classes by training on examples from $\mathcal{D}$. The idea behind the episodic paradigm is to simulate the types of few-shot problems that will be encountered at test, taking advantage of the large quantities of available labeled data from $\mathcal{D}$.

Specifically, models are trained on $K$-way $S$-shot episodes constructed by first sampling a small subset ${V}$ of $K$ classes from $\mathcal{D}$ and then generating: 1) a support set $\mathcal{S}_{V}$ containing $S$ examples from each of the $K$ classes in the sampled subset ${V}$ and 2) a query set $\mathcal{Q}_V$ of different examples from the same $K$ classes. Let $\textsc{RandomSample}(\mathcal{C}, N)$ denote a set of $N$ elements chosen uniformly at random from set $\mathcal{C}$ without replacement. Then, ${V}=\textsc{RandomSample}(\{1, \cdots, L\}, K)$,  $\mathcal{S}_V=\cup_{k=1}^K\mathcal{S}_{V_k}$,  and $\mathcal{Q}_V=\cup_{k=1}^K\mathcal{Q}_{V_k}$, 
where $\mathcal{S}_{V_k}=\textsc{RandomSample}(\mathcal{D}_{V_k}, S)$,  $\mathcal{Q}_{V_k}=\textsc{RandomSample}(\mathcal{D}_{V_k} \setminus \mathcal{S}_{V_k}, Q)$, and $Q$ is the number of query examples in each of the $K$ classes. Training on such episodes is done by feeding the support set $\mathcal{S}_V$ to the model and updating its parameters to minimize the loss of its predictions for the examples in the query set $\mathcal{Q}_V$.

%%\vspace{0.2cm}
\noindent{\bf Prototypical networks}. 
Prototypical Network~\cite{snell2017prototypical} is a few-shot learning model that has the virtue of being simple and yet obtaining state-of-the-art performance. At a high-level, it uses the support set $\mathcal{S}_V$ to extract a prototype vector from each class, and classifies the inputs in the query set $\mathcal{Q}_V$ based on their distance to the prototype of each class.

More precisely, Prototypical Networks learn an embedding function $f_{\bm \phi} : \mathbb{R}^D \rightarrow \mathbb{R}^M$ with learnable parameters $\bm{\phi}$, that maps examples into a space where examples from the same class are close and those from different classes are far. All parameters of Prototypical Networks lie in the embedding function.

To compute the prototype $\mathbf{c}_{V_k}$ of each class $k$ in the selected subset $V$, a per-class average of the embedded examples is performed:
\begin{equation}
\mathbf{c}_{V_k} = \frac{1}{|\mathcal{S}_{V_k}|} \sum_{(\mathbf{x}_{i'}, y_{i'}) \in \mathcal{S}_{V_k}} f_{\bm{\phi}}(\mathbf{x}_{i'})
\label{eq:prototype}
\end{equation}

These prototypes define a predictor for the class of any new (query) example $\mathbf{x}_i$, which assigns a probability over any class ${V_k}$ based on the distances between $\mathbf{x}_i$ and each prototype $\mathbf{c}_{V_k}$, as follows:
\begin{equation}
p_{\bm \phi}(y = V_k\,|\,\mathcal{S}_{V},\mathbf{x}_i) = \frac{\exp(-d(f_{\bm \phi}(\mathbf{x}_i), \mathbf{c}_{V_k}))}{\sum_{k'} \exp(-d(f_{\bm \phi}(\mathbf{x}_i), \mathbf{c}_{V_{k'}}))},
\label{eq:classdist}
\end{equation}
where $d: \mathbb{R}^M \times \mathbb{R}^M \rightarrow [0, +\infty)$ is a distance function. The loss function used to update Prototypical Network for a query example of a given episode is then simply the negative log-probability of the true class $y_i$:
\begin{equation}
\label{eq:loss-func-pn}
J(f_{\bm \phi};\mathcal{S}_{V},\mathbf{x}_i,y_i) = -\log p_{\bm \phi}(y = y_i\,|\,\mathcal{S}_{V},\mathbf{x}_i)
\end{equation}
Training proceeds by minimizing the average loss for all query examples, iterating over training episodes and performing a gradient descent update for each.

Generalization performance is measured on test set episodes, which contain images from classes in $\mathcal{D}_{test}$ instead of $\mathcal{D}$. For each test episode, we use the predictor produced by the Prototypical Network for the provided support set $\mathcal{S}_V$ to classify each of query input $\mathbf{x}_i$ into the most likely class $\hat{y}=\argmax_k p_{\bm \phi}(y = V_k\,|\,\mathcal{S}_{V},\mathbf{x}_i)$.

\subsection{Formal parallel}\label{sec:definition}
{\bf Supervised learning}. Suppose we consider a supervised learning problem on a labeled training dataset $\mathcal{D}^s=\{(\mathbf{s}_i, y_i)\}_{i=1}^{N}$ with the following objective function:
\beq\label{eq:min_pro}
\min_{\bm{\phi}} \mathcal{L}(f_{\bm{\phi}}):=\sum_i l(f_{\bm{\phi}};\mathbf{s}_i, y_i),
\eeq
where $f_{\bm{\phi}}$ is a specified network with parameters $\bm{\phi}$, and $l(\cdot;\cdot,\cdot)$ is a predefined loss function. 

%%\vspace{0.2cm}
\noindent {\bf Regular classification}.
For a regular classification problem on the above defined training dataset $\mathcal{D} =\{ (\mathbf{x}_i, y_i)\}_{i=1}^{LH}=\cup_{j=1}^L\mathcal{D}_j$, $f_{\bm{\phi}}$ is the classifier to be learned, and a plain example of the loss function $l(f_{\bm{\phi}};\mathbf{x}_i,y_i)$ is cross entropy, which takes the form:
\begin{equation}
\begin{aligned}
l(f_{\bm{\phi}};\mathbf{x}_i,y_i)&=-\log p(y=y_i|\mathbf{x}_i),\\ p(y=j|\mathbf{x}_i)&=\frac{\exp{(-f_{\bm{\phi}}(\mathbf{x}_i)_j)}}{\sum_{j'=1}^L \exp{(-f_{\bm{\phi}}(\mathbf{x}_i)_{j'})}}, 
\end{aligned}
\end{equation}
where $f_{\bm{\phi}}(\mathbf{x}_i)_j$ denotes the $j$-th output of $f_{\bm{\phi}}(\mathbf{x}_i)$. 

%%\vspace{0.2cm}
\noindent {\bf Few-shot classification}. Suppose we consider the $K$-way $S$-shot classification problem on the above training dataset $\mathcal{D}$. Motivated by the episodic training paradigm,
we can define a \textbf{task class} as a class label subset ${V} \in \mathcal{T}$ that contains $K$ indexes of the $L$ classes. 
Then we can define a \textbf{task example} $(\bm{\tau}_{i}, y_i)$ of task class ${V}$ as a pair of a \emph{support} set $\mathcal{S}_{V}$, 
and a \emph{query} example $(\mathbf{x}_{i},y_{i})$ from the corresponding query set $\mathcal{Q}_{V}$.
%can generate several task examples $\mathcal{G}({V}):=\{(\bm{\tau}_{i},y_i)\}$, each of which contains a \emph{support} set, and a \emph{query} example. 
To be specific, each task example $(\bm{\tau}_{i}, y_i)$ can be denoted by $(\bm{\tau}_i=\{\mathcal{S}_V,\mathbf{x}_{i}\}, y_{i})$, where $(\mathbf{x}_{i},y_{i})=\textsc{RandomSample}(\mathcal{Q}_{V}, 1)$. 
Thus, the total number of task classes $|\mathcal{T}|$ is $C_{L}^K$, and each task class ${V}$ will have $|\mathcal{G}({V})|=(C_{H}^S)^KC_K^1(H-S)$ task examples. 
As shown in Figure \ref{fig:def}, we give an instance of task classes and task examples for a $3$-way $1$-shot classification problem on 
a dataset with $4$ classes, and each class has $6$ examples. 
% We also show the comparisons between the few-shot classification problem and the classification problem from the view of supervised learning in Table~\ref{tab:comparison}. 
On top of the above formalism of task classes and task examples, the few-shot classification problem can be also formulated as a supervised learning problem represented in Eq.~\eqref{eq:min_pro} like the regular classification problem. Particularly, the regular supervised classification aims to learn a classifier $f_{\bm{\phi}}$ ($f_{\bm{\phi}} (\mathbf{x})$ can estimate the label of example $\mathbf x$) given \textbf{a large number of examples} $\{(\mathbf{x}_i, y_i)\}$, and the few-shot classification problem aims to learn a tasker (the one that can handle a task) $f_{\bm{\phi}}$ ($f_{\bm{\phi}}(\bm\tau)$ can estimate whether we complete the task $\bm \tau$ or not) given \textbf{a large number of task examples} $\{(\bm {\tau}_i, y_i)\}$. To clarify the formulation of the few-shot classification problem, the loss function $l(f_{\bm{\phi}};\bm{\tau}_i,y_i)$ of Prototypical Networks~\cite{snell2017prototypical} represented in Eq.~\eqref{eq:loss-func-pn} is used for instance, which has a well correspondence with the cross-entropy loss in the regular classification problem given the definition of $\bm{\tau}_i=\{\mathcal{S}_V,\mathbf{x}_{i}\}$.

\iffalse
\begin{equation}
\begin{aligned}
&l(f_{\bm{\phi}};\bm{\tau}_i,y_i)=-\sum_{k=1}^K{\bf 1}(y_i==V_k)\log p(y=V_k|\bm{\tau}_i),\\ &p(y=V_k|\bm{\tau}_i)=\frac{\exp{(-d(f_{\bm{\phi}}(\mathbf{x}_i),\mathbf{c}_{V_k}^i))}}{\sum_{k'=1}^K \exp{(-d(f_{\bm{\phi}}(\mathbf{x}_i),\mathbf{c}_{V_{k'}}^i))}},\\
&\mathbf{c}_{V_k}^i=\frac{1}{|\mathcal{S}_{V_k}^i|}\sum_{(\mathbf{x}_{i'},y_{i'})\in \mathcal{S}_{V_k}^i} f_{\bm{\phi}}(\mathbf{x}_{i'})\\
\end{aligned}
\end{equation}
\fi

\begin{figure}[!htb]
	%\vspace{-0.3cm}
	\centering
	\subfigure[One-episode]         % Label of subfigure in {}
	{\includegraphics[width=0.2\textwidth ]{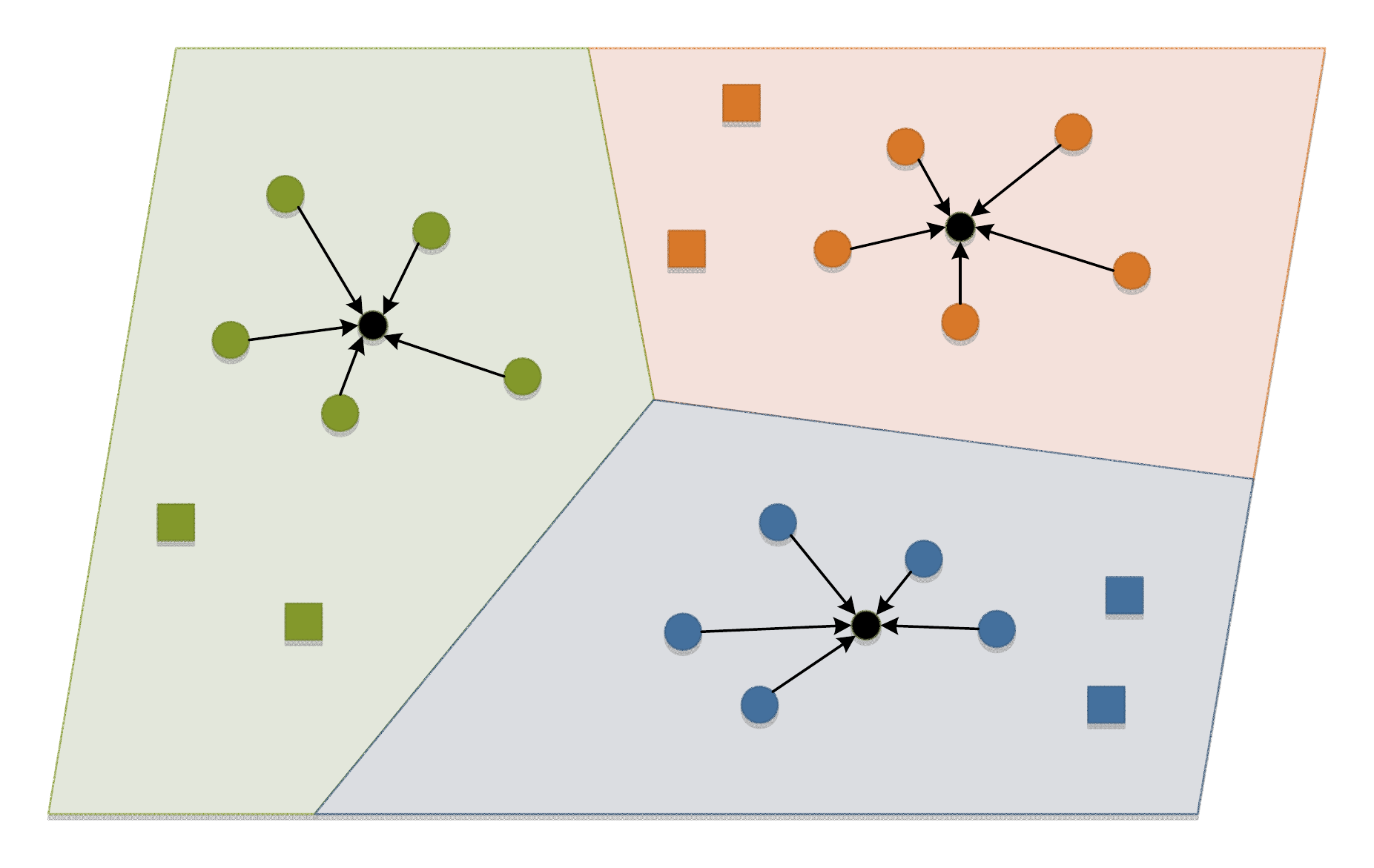}
		\label{fig:multi-episode-training:one-episode}}
	\subfigure[Multi-episode]         % Label of subfigure in {}
	{\includegraphics[width=0.35\textwidth ]{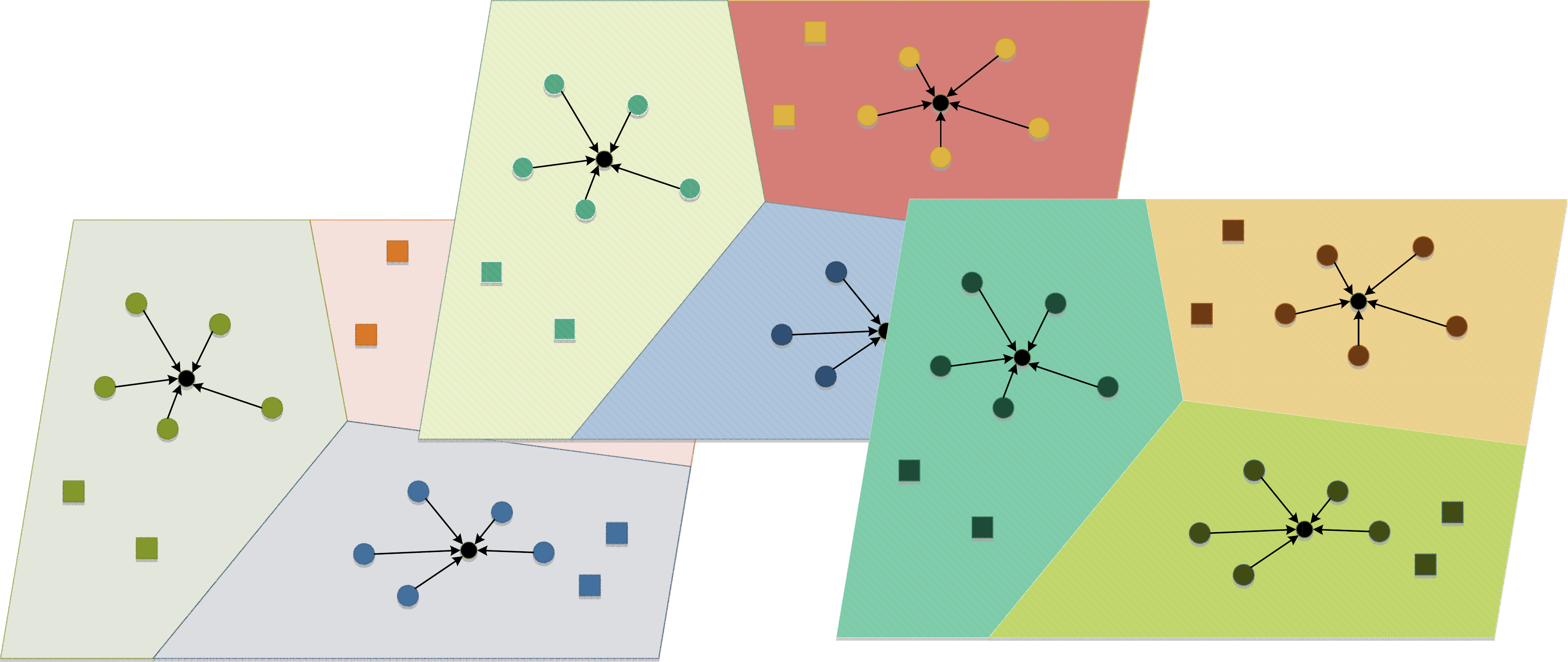}
		\label{fig:multi-episode-training:multi-episode}}
	%%\vspace{-0.2cm}
	\caption{A illustration comparison between the one-episode and multi-episode training strategy for a 3-way 5-shot problem at each iteration. Support and query examples are shaped in circle and square respectively. } 
	\label{fig:multi-episode-training}
	%%\vspace{-0.3cm}
\end{figure}

\subsection{Minibatch training}
\label{sec:minibatch}
We discuss the minibatch training from the view of stochastic gradient descent (SGD)~\cite{Robbins1951A}. 
To solve the supervised learning problem in Eq.~\eqref{eq:min_pro}, minibatch SGD performs the following update:
\beq\label{eq:mini-batch-SGD}
\bm{\phi}_{t+1}=\bm{\phi}_t-{\alpha}\sum_{(\mathbf{s}_i, y_i)  \in \mathcal{B}_t} \frac{\partial l(f_{\bm{\phi}};\mathbf{s}_i,y_i)}{\partial \bm{\phi}}\, |_{\bm{\phi}=\bm{\phi}_t},
\eeq
where $\alpha$ is the learning rate, $t$ is the iteration index, and $\mathcal{B}_t$ is a minibatch randomly sampled from the whole dataset $\mathcal{D}^s$. 
Minibatch SGD is proven to be both efficient and effective~\cite{Dekel2012Optimal}. 

%%\vspace{0.2cm}
\noindent {\bf Regular classification}. 
Formally, for a regular classification problem on the training dataset $\mathcal{D}$, at each training step represented by Eq.~\eqref{eq:mini-batch-SGD}, we randomly select several examples in $\mathcal{D}$ uniformly as $\mathcal{B}_t$. 
Suppose $|\mathcal{B}_t|=100$ and $L=10$; then, from the perspective of probability, 
$\mathcal{B}_t$ will have approximately $10$ examples from each $\mathcal{D}_j$. 

%%\vspace{0.2cm}
\noindent {\bf Few-shot classification}. For a $K$-way $S$-shot classification problem with $|\mathcal{T}| = C_{L}^K$ task classes and $|\mathcal{G}({V})|=(C_{H}^S)^KC_K^1(H-S)$ task examples in each task class $V \in \mathcal{T}$, 
we have the dataset denoted as 
$\mathcal{D}^f =\{(\bm{\tau}_i, y_i)\}_{i=1}^{|\mathcal{T}||\mathcal{G}({V})|} =\cup_{V \in \mathcal{T}}\, \mathcal{G}({V})$, 
where $\bm{\tau}_i=\{\mathcal{S}_V,\mathbf{x}_{i}\}$, $(\mathbf{x}_{i},y_{i})=\textsc{RandomSample}(\mathcal{Q}_{V}, 1)$. 
Here, the minibatch training for few-shot classification is that at each training step represented by Eq.~\eqref{eq:mini-batch-SGD}, we should randomly select several task examples in $\mathcal{D}^f$ as $\mathcal{B}_t$. 
Apparently, the number of the whole set of task examples of $\mathcal{D}^f$ is extremely massive, as that calculated in the above section~\ref{sec:definition}. 
Thus it is almost impossible to explicitly generate $\mathcal{D}^f$\footnote{Intuitively, we should explicitly generate the whole set of task examples of $\mathcal{D}^f$ and then sample a minibatch in $\mathcal{D}^f$ uniformly.}, which is definitely memory- and time- consuming. 
Here, we recommend \textbf{multi-episode training}: sampling several task classes in $\mathcal{T}$ and then sampling several task examples of the sampled task classes as a minibatch. 
From this point of view, the episodic training paradigm proposed by matching nets~\cite{vinyals2016matching}, which samples an episode that consists of a pair of a support set $\mathcal{S}_{V}$ and a query set $\mathcal{Q}_V$ at each training iteration, can be seen as 
randomly selecting \emph{only one} task class $V$ in $\mathcal{T}$ and then sampling \emph{$KQ$ task examples of the same support set $\mathcal{S}_{V}$ in the selected task class $V$} as $\mathcal{B}_t$, namely $\mathcal{B}_t \subset \mathcal{G}(V)$. 
Obviously, it is not a reasonable choice because it is almost impossible that the task examples in $\mathcal{B}_t$ just located in one task class given $\mathcal{B}_t$ are exactly randomly sampled from $\mathcal{D}^f$ uniformly. 
Multi-episode training is proposed to relieve this issue by using multiple episodes to construct $\mathcal{B}_t$. 
Denote $E$-episode training such that we use minibatch $\mathcal{B}_t:= \cup_{e=1}^E \mathcal{B}_t^e$, where $\mathcal{B}_t^e \subset \mathcal{G}(V^e)$ is a randomly sampled episode, and $V^e$, $e=1,\cdots,E$ are $E$ task classes first randomly sampled from $\mathcal{T}$. Figure \ref{fig:multi-episode-training} shows a comparison instance between the one-episode and multi-episode training strategy for a 3-way 5-shot problem at each iteration.
In addition, one may instead use randomly sampled task examples (of different support sets) and not use a randomly sampled episode in each task class to construct minibatch $\mathcal{B}_t$, which deserves deeper experimental investigations at future\footnote{We should notice that the number of task examples in each task class is still extremely massive, which makes the uniform sampling at each task class remain complicated.}.

%%\vspace{0.2cm}
\noindent {\bf Relation with MAML~\cite{FinnAL2017Model}}.
The idea of multi-episode training has already been used in optimization based few-shot classification methods, such as MAML~\cite{FinnAL2017Model}, but with different motivation. In MAML, sampling multiple episodes aimes to optimize the performance across different kinds of tasks, while in this paper, the idea is directly motivated by the minibatch training in regular classification problems and aims to reduce the number of iterations before convergence.

\begin{figure}[!htb]
	%%\vspace{-0.3cm}
	\centering
	\includegraphics[width=0.6\textwidth]{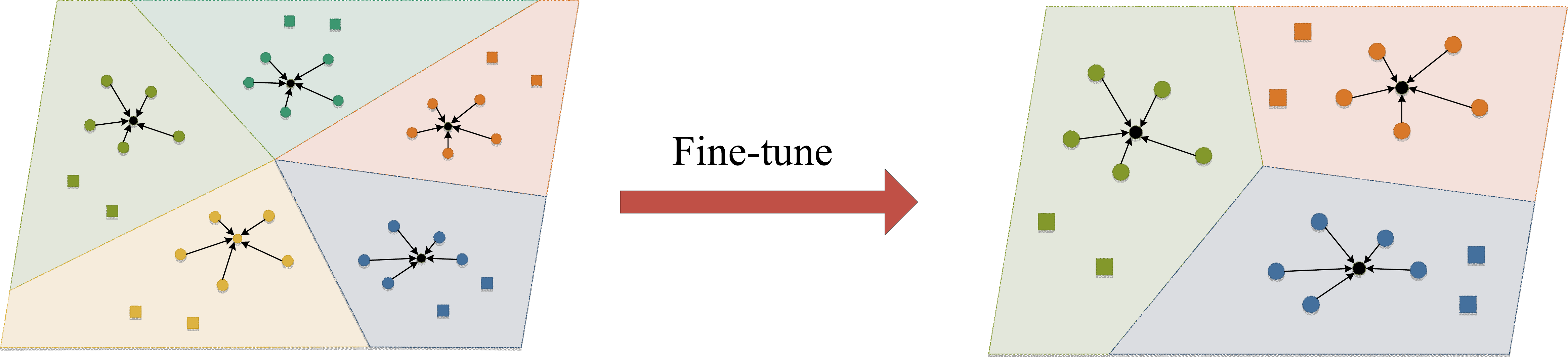}
	%%\vspace{-0.2cm}
	\caption{A illustration of the cross-way training strategy for a 3-way 5-shot problem, pretrained by a 5-way 5-shot problem. Support and query examples are shaped in circle and square respectively.} 
	\label{fig:cross-way-training}
	%%\vspace{-0.5cm}
\end{figure}

\subsection{Pretraining}
Another critical point of iteratively solving the supervised learning problem represented by Eq.~\eqref{eq:min_pro} is the initial value, $\bm{\phi}_0$ in Eq.~\eqref{eq:mini-batch-SGD}. 
The pretraining strategy suggests a choice of $\bm{\phi}_0$, which comes from solving another supervised learning problem with a similar or more complicated data distribution. 
Specifically, suppose we have another data distribution, $\mathcal{D}_{\rm pre}$. Then, we set the initial value $\bm{\phi}_0$ for solving Eq.~\eqref{eq:min_pro} as:
\begin{equation}
\bm{\phi}_0={\rm argmin}_{\bm{\phi}} \, \sum_{(\mathbf{s}_i, y_i) \in \mathcal{D}_{\rm pre}} l(f_{\bm{\phi}};\mathbf{s}_i, {y}_i).
\end{equation}

\noindent {\bf Regular classification}. We usually set $\mathcal{D}_{\rm pre}$ to be a dataset of large-scale data, that is, $|\mathcal{D}| < |\mathcal{D}_{\rm pre}|$. 
The most famous pretraining approach is ImageNet pretraining~\cite{Russakovsky2015ImageNet} for computer vision tasks. 
ImageNet pretraining has achieved state-of-the-art results on many computer vision tasks, such as object detection~\cite{ren2015faster} and image segmentation~\cite{he2017mask}. 
Recently, ImageNet pretraining has been found to speed up the convergence but not necessarily improve the final accuracy~\cite{he2018rethinking}. 

%%\vspace{0.2cm}
\noindent {\bf Few-shot classification}. Similar to using large-scale data to pretrain in classification problems, here we suggest using the $\hat K$-way $S$-shot classification problem to pretrain the $K$-way $S$-shot classification problem with $K < \hat K < L-K$, because obviously we have the number of task classes $C_{L}^K < C_{L}^{\hat K}$ and the total number of task examples 
$C_{L}^K (C_{H}^S)^KC_K^1(H-S)< C_{L}^{\hat K} (C_{H}^S)^{\hat K}C_{\hat K}^1(H-S)$. We named this kind of pretraining for few-shot classification the \textbf{cross-way training} strategy\footnote{It should be noted that the network in pretraining does not need to be the same as that in the target supervised learning problem. Therefore, in addition to metric-based methods, the cross-way training strategy can also work for optimization or memory-based methods, where we can optionally just pretrain several feature layers of the network. Related work can be explored in the future.}. Figure~\ref{fig:cross-way-training} shows a cross-way training strategy instance for a 3-way 5-shot problem, pretrained by a 5-way 5-shot problem. 
\noindent {\bf Relation with the training idea in Prototypical Networks~\cite{snell2017prototypical}}.
Prototypical Networks also suggested to train on a different "way", but they do not explore its actual impact and not expand it with fine-tuning. Particularly, different from the conclusion in ~\cite{snell2017prototypical}, empirical evaluation conducted in this paper verify that pretraining employing a higher way without fine-tuning cannot always achieve a better test accuracy on the target problem with a lower way than training from random initialization if given a slower learning rate decaying policy and more iteration steps\footnote{We also find that the improvement of fine-tuning is marginal, which may occur because the task examples of $\hat K$-way or $K$-way are generated from the same original dataset. In the future, we may investigate the exact pretraining strategy for few-shot classification problems by pretraining few-shot classification problems based on different original datasets.}.

\vspace{-0.3cm}
\section{Experiments}
\noindent {\bf Baselines}. We consider testing the performance of the proposed training strategies on Prototypical Networks~\cite{snell2017prototypical}. 
The basic embedding architecture used in Prototypical Networks is composed of four
convolutional blocks. Each block comprises a $64$-filter $3\times 3$ convolution, batch normalization layer~\cite{IoffeS2015Batch}, a ReLU nonlinearity and a $2\times 2$ max-pooling layer. 
To simplify the discussion, we refer to the detailed description of net architectures in their papers. In our experiments, both of the proposed training strategies are implemented based on the open source code of Prototypical Networks~\footnote{https://github.com/jakesnell/prototypical-networks.git}. 

\noindent {\bf Datasets}. We performed experiments on the two related datasets: Omniglot~\cite{Lake2011One} and the \emph{mini}ImageNet version of ILSVRC-2012~\cite{Russakovsky2015ImageNet} with the splits proposed by Ravi and Larochelle~\cite{ravi2017optimization}. 
%Omniglot is a dataset of $1623$ handwritten characters collected from $50$ alphabets.  Each character contains $20$ examples, each of which is drawn by a different human subject. 
For Omniglot, we follow the procedure of Vinyals et al.~\cite{vinyals2016matching} by resizing the grayscale images to $28\times 28$ and augmenting the character classes with rotations in multiples of 90 degrees. We use 1028 characters plus rotations for training (4,112 classes in total), 172 characters plus rotations for validation (688 classes in total) and the remaining 1692 classes including rotations for testing. We follow the procedure of %Snell et al.~
\cite{snell2017prototypical} by first training on the 4112 training classes and
using the 688 validation classes for monitoring generalization performance and finding the optimal number of iteration ($T$, after which validation loss
stops improving) and then training on both the 4112 training classes and the 688 validation classes with $T$ iterations. 
The \emph{mini}ImageNet dataset, originally proposed by~\cite{vinyals2016matching}, consists of $60,000$ color images % of size $84 \times 84$ 
divided into 100 classes with 600 examples each. 
%For \emph{mini}ImageNet, 
We use the splits introduced by~\cite{ravi2017optimization}. 
Their splits use a different set of 100 classes, divided into 64 training, 16
validation, and 20 testing classes. We follow their procedure by training on the 64 training classes and using the 16 validation classes for monitoring generalization performance only.

\noindent {\bf Training settings}. 
Following Prototypical Networks, we consider the 1-shot and 5-shot scenarios, 5-way and 20-way situations under Euclidean distance for Omniglot, and 5 way only under cosine\footnote{As discussed in~\cite{snell2017prototypical}, Prototypical Networks with cosine distance are equivalent to matching networks~\cite{vinyals2016matching} in the case of one-shot learning. Therefore, these results also demonstrate the applicability of the proposed training strategies to matching networks.} and
Euclidean distance for \emph{mini}ImageNet. 
We consider multi-episode training with the number of episodes $E=1,3,5$. 
The maximum number of iterations for $E=1,3,5$ are set as 450000, 150000, 90000, respectively, to make sure all models are trained with the same size of data. 
In each episode, we set the number of query points per class in the selected task class as 15%\footnote{This implies there are $15 \times 5 + 1 \times 5 = 80$ images in one episode for the 5-way 1-shot problem.}
. Regarding cross-way training, we fixedly set the number of episodes in each iteration as $1$. We then consider training with a higher way of $60$ for Omniglot and set the number of query points per class as $5$. For \emph{mini}ImageNet, we consider training with higher ways of $20, 30$ and set the number of query points per class as $15$. It should be noted that for \emph{mini}ImageNet, we can only monitor the 5-way validation performance during higher-way training because there are only 16 classes in the validation set. The maximum number of iterations for fine-tuning the pretrained models is set as $20000$. All of the models were trained via SGD with Adam~\cite{kingma2014adam}. We used an initial learning rate of $10^{-3}$. The original Prototypical Networks cut the learning rate in half every $2000$ iterations (namely,``$1\times$ schedule''). For models in this paper, we investigate the slower learning rate scheduling strategy\footnote{Detailed results of different learning rate scheduling strategy are included in the supplementary material.}, and
we use a similar terminology, e.g., a so-called ``$3\times$ ($5\times$) schedule''
to cut the learning rate in half every $6000$ ($10000$) iterations. 
%We find that training longer for large learning rates is useful for one-episode or lower-way training, but training longer at small learning rates often leads to underfitting.  
No regularization was used other than batch normalization. All our models are end-to-end trained from scratch with no additional dataset. 

\noindent {\bf Testing settings}. For all tasks considered in our experiments, we set the number of test points per class in each testing episode as 15. We computed few-shot classification accuracies for our models on Omniglot (\emph{mini}ImageNet) by averaging over $1000$ ($600$) randomly generated episodes from the testing set with $95\%$ confidence intervals. 
\vspace{-0.2in}
\begin{table*}[!htb]
	\caption{The optimum convergence accuracy-iterations results of multi-episode training on Prototypical Networks~\protect\cite{snell2017prototypical} for Omniglot. * Results reported in the original paper~\protect\cite{snell2017prototypical}. } \label{tab:multi-episode}
	\centering
	%\vspace{-0.05in}
	\begin{small}
		\begin{spacing}{1.2}
			\scalebox{0.72}[0.72]{
				
				\begin{tabular}{l"cc|cc"cc|cc}
					\specialrule{1pt}{0pt}{2pt}		
					\multicolumn{1}{c"}{\multirow{3}{*}{{$ E$}}}&	\multicolumn{4}{c"}{\textbf{\textsc{5 Way }}}&\multicolumn{4}{c}{\textbf{\textsc{20 way}}}\\
					
					&\multicolumn{2}{c|}{{\textsc{ 1 shot}}}&\multicolumn{2}{c"}{{\textsc{5 shot}}}	&\multicolumn{2}{c|}{{\textsc{ 1 shot}}}&\multicolumn{2}{c}{{\textsc{5 shot}}}\\
					
					&\textit{Acc.} & \textit{Iters.($\scriptstyle 10^2$)} &\textit{Acc.} & \textit{Iters.($\scriptstyle 10^2$)} &\textit{Acc.} & \textit{Iters.($\scriptstyle 10^2$)}  &\textit{Acc.} & \textit{Iters.($\scriptstyle 10^2$)} \\ 
					\thickhline
					$\bm 1^*$&	97.40\%          & -    &99.30\%  & - 
					&	95.40\%          & -   &	98.70\%  &-  \\ 
					\hline
					$\bm 1$& {98.28 $\pm$ 0.19\%}&	{1094}&{99.53 $\pm$ 0.07\%}&	{935}&{95.10 $\pm$ 0.17\%}&	{506}&	{98.70 $\pm$ 0.06\%}&	{506}\\
					$\bm 3$&	{98.49 $\pm$ 0.17\%}&	{812}&{99.59 $\pm$ 0.07\%}&	\textbf{440}&\textbf{95.11 $\pm$ 0.17\%}&	{477}&	\textbf{98.75 $\pm$ 0.06\%}&	{519}\\
					$\bm 5$&	\textbf{98.56 $\pm$ 0.16\%}&	\textbf{517}&\textbf{99.61 $\pm$ 0.07\%}&	{760}&	{94.98 $\pm$ 0.17\%}&	\textbf{468}&{98.70 $\pm$ 0.06\%}&	\textbf{265}\\			
					\specialrule{1pt}{2pt}{0pt}
				\end{tabular}
				
			}
		\end{spacing}
	\end{small}
\end{table*}

\begin{table}[!htb]
	\caption{The optimum convergence accuracy-iterations results of multi-episode training on Prototypical Networks~\protect\cite{snell2017prototypical} for \emph{mini}ImageNet. 
		*Results reported in the original paper~\protect\cite{snell2017prototypical}. } \label{tab:multi-episode-imagenet}
	%\vspace{-0.05in}
	\centering
	\begin{small}
		\begin{spacing}{1.2}
			\scalebox{0.78}[0.78]{
				\begin{tabular}{l"l"cc|cc}
					\specialrule{1pt}{0pt}{2pt}
					
					\multicolumn{1}{l"}{\multirow{3}{*}{\textbf{Dist.}}}&{\multirow{3}{*}{{$E$}}}& \multicolumn{4}{c}{\textbf{\textsc{5 Way}}}\\
					&&\multicolumn{2}{c|}{{\textsc{1 shot}}}&\multicolumn{2}{c}{{\textsc{5 shot}}}\\
					&&	\textit{Acc.} & \textit{Iters.($\scriptstyle 10^2$)}&	\textit{Acc.} & \textit{Iters.($\scriptstyle 10^2$)}\\
					\thickhline
					\multicolumn{1}{l"}{\multirow{4}{*}{{Cosine}}}&$\bm 1^*$ &	38.82 $\pm$ 0.69\%          & -& 	51.23 $\pm$0.63\%  & -\\
					\cline{2-6}	
					&$\bm 1$&	{41.08 $\pm$ 0.72\%}&	{837}&	{49.78 $\pm$ 0.67\%}&	{222}\\
					&$\bm 3$&{41.47 $\pm$ 0.71\%}	&\textbf{216}&\textbf{49.93 $\pm$ 0.68\%}&	{158}\\
					&$\bm 5$&\textbf{41.54 $\pm$ 0.73\%}&	{247}&	{49.87 $\pm$ 0.70\%}&	\textbf{94}\\	
					\thickhline
					\multicolumn{1}{l"}{\multirow{4}{*}{{Euclid.}}}&$\bm 1^*$ &	46.61 $\pm$ 0.78\%          & -& 	65.77 $\pm$ 0.70\%  & -\\
					\cline{2-6}
					&$\bm 1$&	{48.39 $\pm$ 0.80\%}&	{483}&	\textbf{66.24 $\pm$ 0.65\%}&	{458}\\
					&$\bm 3$&{48.81 $\pm$ 0.79\%}&	{301}&{65.85 $\pm$ 0.65\%}&	\textbf{122}\\
					&$\bm 5$&\textbf{49.00 $\pm$ 0.80\%}&	\textbf{225}&{65.77 $\pm$ 0.67\%}&	{273}\\	
					\specialrule{1pt}{2pt}{0pt}
				\end{tabular}
			}
		\end{spacing}
	\end{small}
\end{table}

\vspace{-0.5in}
\subsection{Multi-Episode Training}
\label{sec:exp-multi-episode}
\begin{figure}[!htb]
	\centering
	\includegraphics[width=0.35\textwidth]{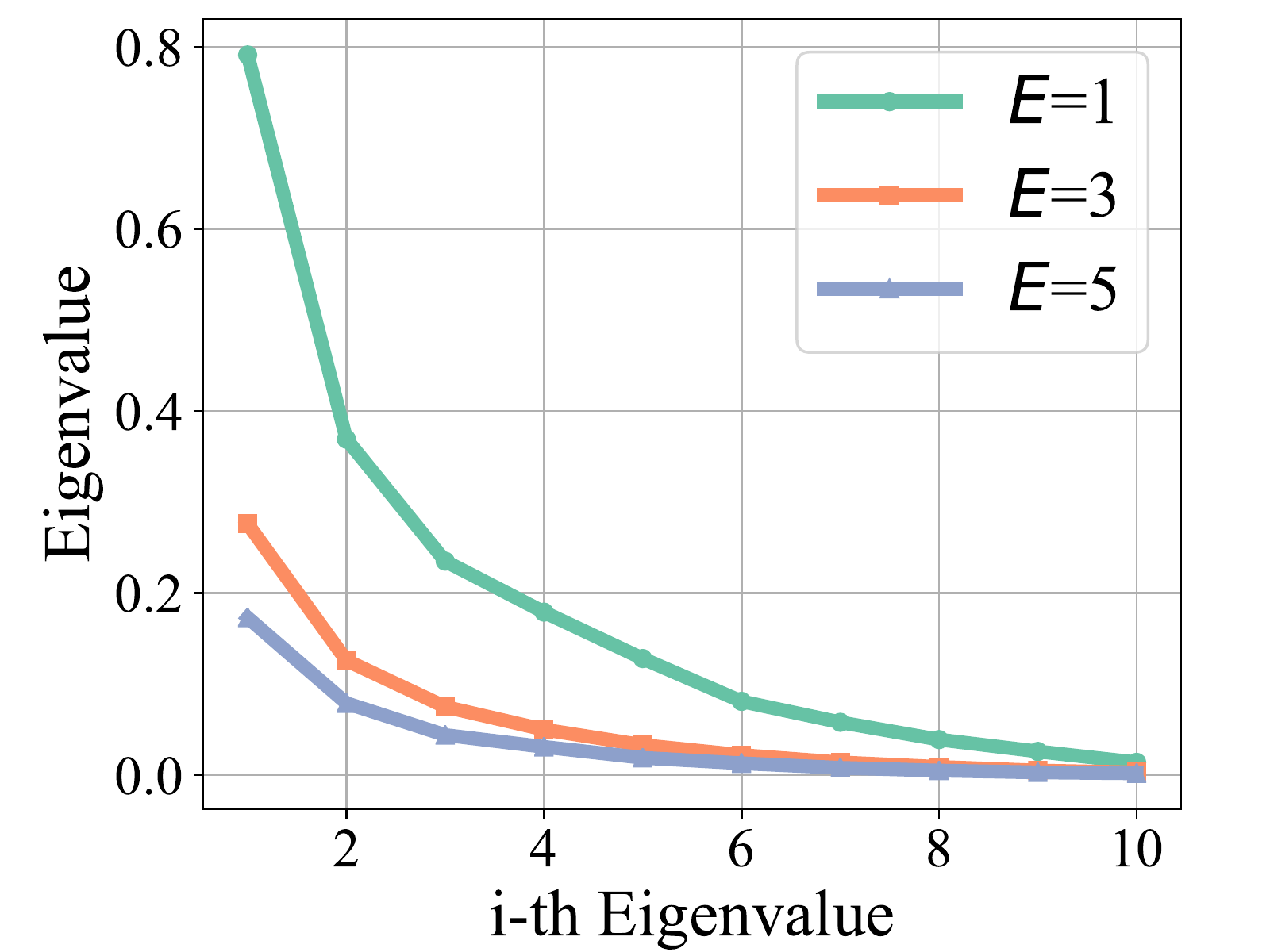}
	\includegraphics[width=0.35\textwidth]{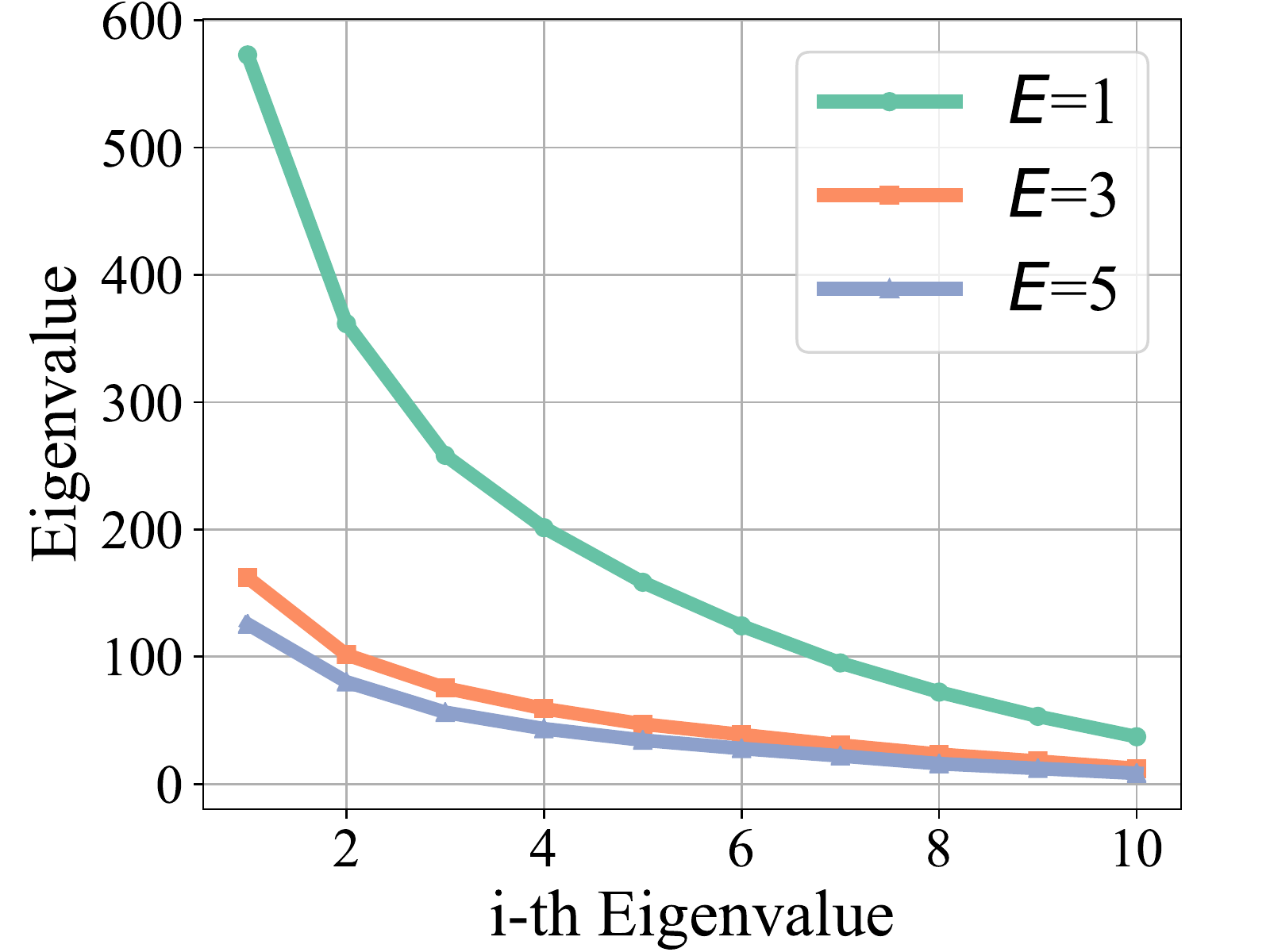}
	%\vspace{-0.1in}
	\caption{Top 10 eigenvalues of the Hessian of the loss function w.r.t weights is shown for multi-episode training on Prototypical Networks (with Euclidean distance) for Omniglot (left) and \emph{mini}ImageNet (right) 5 way 1 shot task. The spectrum is computed using power iteration~\cite{Lee2017FirstorderMA} with relative error of 1E-4. This figure shows that multi-episode training converges to points with noticeably smaller Hessian spectrum compared with one-episode training. As discussed in~\cite{yao2018hessianbased},  points with small or flat Hessian spectrum show robustness to adversarial perturbation and yield accuracy improvement.}
	%\vspace{-0.1in}
	\label{fig:hessianeigenvalue}
\end{figure}

%\vspace{-0.05in}
The optimum convergence accuracy-iterations results of multi-episode training with different episode numbers for Omniglot and \emph{mini}ImageNet are shown in Table ~\ref{tab:multi-episode} and Table ~\ref{tab:multi-episode-imagenet} respectively. The rows where the episode number is set as $1^*$ and $1$ report the results presented in~\cite{snell2017prototypical}, and we rerun the open source code the paper provided, respectively\footnote{\label{footnote:resultsexplain}Since the open source code does not contain the implementations of training settings and data processing for \emph{mini}ImageNet, we have to set these related hyperparameters or re-implement the data processing code ourselves. Therefore, results are slightly different on \emph{mini}ImageNet.}. These results demonstrate that:
\begin{itemize} 
	\item With the increasing of the number of episodes in each iteration, models can converge faster for most considered tasks. Similar phenomena are consistently
	present for both Omniglot and \emph{mini}ImageNet. Specifically, for Omniglot 5-way 1-shot, 20-way 5-shot tasks and \emph{mini}ImageNet 5-way 1-shot tasks, the iteration number of the multi-episode training with $5$ episodes reduces nearly or more than half compared with only $1$ episode. It should be noticed that the optimal number of episodes are not consistent for different tasks. This result is in line with what usually happens in regular supervised learning~\cite{Dekel2012Optimal}.
	\item Regarding the accuracy performance, models trained with multiple episodes in each iteration appear to be superior to those trained with one episode on most tasks. As shown in Table~\ref{tab:multi-episode}, for Omniglot 5-way 1-shot, 5-way 5-shot tasks and \emph{mini}ImageNet 5-way 1-shot tasks, the accuracy of models improves with the increase in episode numbers in each iteration, namely, 5 episodes are better than 3 episodes, and 3 episodes are better than 1 episode. Specifically, for the Omniglot $5$-way $1$-shot problem, the multi-episode training with $5$ episodes can achieve approximately $0.28\%$ accuracy improvement compared with only $1$ episode on a very high baseline accuracy of $98.28\%$. 
\end{itemize}
In total, training with more episodes helps to speed up convergence on the target task and improves the final classification accuracy on most tasks. 
This may because that multi-episode training can well relieve the problem of imbalanced task class sampling of one-episode training, or it may improve the behaviour of Hessian spectrum and make the optimization more robust~\cite{yao2018hessianbased}, see Figure~\ref{fig:hessianeigenvalue}. 

\begin{table*}[!htb]
	\vspace{-0.2in}
	\caption{The optimum convergence accuracy-iterations results of cross-way training on Prototypical Networks~\protect\cite{snell2017prototypical} for Omniglot. FT is short for fine-tune.
		*Results reported the original paper~\protect\cite{snell2017prototypical}. } \label{tab:cross-way}
	%\vspace{-0.2in}
	\begin{center}
		\begin{small}
			\begin{spacing}{1.2}
				\scalebox{0.58}[0.58]{	
					\begin{tabular}{l"c|cc|c|cc"c|cc|c|cc}	
						\specialrule{1pt}{0pt}{2pt}
						\multicolumn{1}{l"}{\multirow{3}{*}{$\hat K$}}	&\multicolumn{6}{c"}{\textbf{\textsc{5 Way}}($K=5$)}& \multicolumn{6}{c}{\textbf{\textsc{20 Way}}($K=20$)}\\ 
						
						&\multicolumn{3}{c|}{{\textsc{1 shot}}}  &\multicolumn{3}{c"}{{\textsc{5 shot}}} 	&\multicolumn{3}{c|}{{\textsc{1 shot}}}  &\multicolumn{3}{c}{{\textsc{5 shot}}}                         \\

						& \multicolumn{1}{c|}{\textit{Acc.*}} 	& \multicolumn{1}{c}{\textit{Acc.}} & \multicolumn{1}{c|}{\multirow{1}{*}{\textit{Iters.($\scriptstyle 10^2$)}}} 	&  \multicolumn{1}{c|}{\textit{Acc.*}} 	& \multicolumn{1}{c}{\textit{Acc.}} & \multicolumn{1}{c"}{\multirow{1}{*}{\textit{Iters.($\scriptstyle 10^2$)}}} 		& \multicolumn{1}{c|}{\textit{Acc.*}} 	& \multicolumn{1}{c}{\textit{Acc.}} & \multicolumn{1}{c|}{\multirow{1}{*}{\textit{Iters.($\scriptstyle 10^2$)}}} 		& \multicolumn{1}{c|}{\textit{Acc.*}} 	& \multicolumn{1}{c}{\textit{Acc.}} & \multicolumn{1}{c}{\multirow{1}{*}{\textit{Iters.($\scriptstyle 10^2$)}}} 	 	 \\ 
						
						%\specialrule{0.5pt}{2pt}{2pt}
						\thickhline
						
						$\bm 5$  &97.4\% &	{98.28 $\pm$ 0.19\%}&	{1094} &99.3\% &{99.53 $\pm$ 0.07\%}&	{935}& 92.0\%&	{93.89 $\pm$ 0.19\%}&	{1094}& 97.8\%&	{98.36 $\pm$ 0.06\%}&	{935}\\
						
						$\bm 20$ &98.7\%&{98.70 $\pm$ 0.16\%}&	{506}&99.6\%&{99.63 $\pm$ 0.07\%}&	{506}&95.4\% &	{95.10 $\pm$ 0.17\%}&	{506}& 98.7\% &	{98.70 $\pm$ 0.06\%}&	{506}\\

						$\bm 20$-FT &-&98.45 $\pm$ 0.17\%&	{121}&-&99.56 $\pm$ 0.08\%&	{157}&-&-	&	-& -&	-&	-\\
						$\bm 60$ &98.8\%&\textbf{98.73 $\pm$ 0.15\%}&	{326}&99.7\%&\textbf{99.64 $\pm$ 0.07\%}&{402}&96.0\%
						&	{95.21 $\pm$ 0.17\%}&	{326}& 98.9\%&\textbf{98.77 $\pm$ 0.05\%}&	{402}\\
						$\bm 60$-FT &-&98.45 $\pm$ 0.17\%&	\textbf{112}&-&99.57 $\pm$ 0.07\%&	\textbf{144} &-&	\textbf{95.24 $\pm$ 0.17\%} &	\textbf{191}& -&98.72 $\pm$ 0.06\%&	\textbf{176}\\
						
						\specialrule{1pt}{2pt}{0pt}
				\end{tabular} }
			\end{spacing}
		\end{small}
	\end{center}
	\vspace{-0.4in}
\end{table*}

\begin{table*}[!htb]
	%\vspace{-0.2in}
	\caption{The optimum convergence accuracy-iterations results of cross-way training on Prototypical Networks~\protect\cite{snell2017prototypical} for \emph{mini}ImageNet. FT is short for fine-tune.
		*Results reported in the original paper~\protect\cite{snell2017prototypical}. } \label{tab:cross-way-imagenet}
	%\vspace{-0.2in}
	\begin{center}
		\begin{small}
			\begin{spacing}{1.2}
				\scalebox{0.7}[0.7]{	
					\begin{tabular}{l"l"c|cc"c|cc}	
						\specialrule{1pt}{0pt}{2pt}
						\multicolumn{1}{l"}{\multirow{3}{*}{\textbf{Dist.}}}&	{\multirow{3}{*}{$\hat K$}}&\multicolumn{6}{c}{\textbf{\textsc{5 Way}}($K=5$)}\\
						&	&\multicolumn{3}{c"}{{\textsc{1 shot}}}  &\multicolumn{3}{c}{{\textsc{5 shot}}}                        \\

						&& \multicolumn{1}{c|}{\textit{Acc.*}} 	& \multicolumn{1}{c}{\textit{Acc.}} & \multicolumn{1}{c"}{\multirow{1}{*}{\textit{Iters.($\scriptstyle 10^2$)}}} 	& \multicolumn{1}{c|}{\textit{Acc.*}} 	& \multicolumn{1}{c}{\textit{Acc.}} & \multicolumn{1}{c}{\multirow{1}{*}{\textit{Iters.($\scriptstyle 10^2$)}}}        	 \\ 
						%\specialrule{0.5pt}{2pt}{2pt}
						\thickhline
						\multicolumn{1}{l"}{\multirow{5}{*}{{Cosine}}}& $\bm 5$  & 38.82  $\pm$ 0.69\%& 	41.08 $\pm$  0.72\%&	{837}& 51.23 $\pm$ 0.63\%&	{49.78 $\pm$ 0.67\%}&	{222}\\
						&$\bm 20$ & 43.63 $\pm$ 0.76\%&	{42.03 $\pm$ 0.76\%}&	{151}& 51.48 $\pm$ 0.70\%&	{50.60 $\pm$ 0.68\%}&	{153}\\
						&$\bm 20$-FT & -&	42.28 $\pm$  0.74\%&	{194}& -&		\textbf{52.18 $\pm$  0.70\%}&	{134}\\
						&$\bm 30$ &- &\textbf{42.99 $\pm$ 0.78\%}&	\textbf{138}	& - &	{50.34 $\pm$ 0.68\%}&	\textbf{43}\\
						&$\bm 30$-FT & - &	42.35 $\pm$  0.76\%&	{194}	& - &	52.00 $\pm$  0.68\%&	{110}\\
						%\specialrule{0.5pt}{2pt}{2pt}
						\thickhline
						\multicolumn{1}{l"}{\multirow{5}{*}{{Euclid.}}}& $\bm 5$  & {46.61 $\pm$ 0.78\%}&	{{48.39 $\pm$ 0.79\%}}&	{483}& {65.77 $\pm$ 0.70\%}&	{{66.24 $\pm$ 0.65\%}}&	{458}\\
						&$\bm 20$ & {48.57 $\pm$ 0.79\%}&	{{48.20 $\pm$ 0.80\%}}&	{153}& {68.20 $\pm$ 0.66\%}&	{{65.68 $\pm$ 0.66\%}}&	{453}\\
						&$\bm 20$-FT & -&	{\textbf{49.35 $\pm$ 0.82\%}}&	{144}& -&	{\textbf{66.88 $\pm$ 0.66\%}}&	{150}\\
						&$\bm 30$ &{49.42 $\pm$ 0.78\%} &{{48.37 $\pm$ 0.77\%}}&	{138}	& {66.79 $\pm$ 0.66\%} &	{{65.24 $\pm$ 0.69\%}}&	\textbf{63}\\
						&$\bm 30$-FT & - &	{{49.20 $\pm$ 0.80\%}}&	\textbf{91}	& - &	{{66.41 $\pm$ 0.65\%}}&	{102}\\
						\specialrule{1pt}{0pt}{2pt} 			
				\end{tabular} }
			\end{spacing}
		\end{small}
	\end{center}
	\vspace{-0.5in}
\end{table*}

\subsection{Cross-way Training}
The optimum convergence accuracy-iterations results of cross-way training for Omniglot and \emph{mini}ImageNet are shown in Table~\ref{tab:cross-way} and Table~\ref{tab:cross-way-imagenet} respectively. *Results are presented in the original paper~\cite{snell2017prototypical}. Due to the same reason we discussed in footnote~\ref{footnote:resultsexplain}, our rerunning results on \emph{mini}ImageNet are slightly lower than that provided in the original paper. These results demonstrate the following:
\begin{itemize}
	\item Different from the conclusion in the original paper~\cite{snell2017prototypical}, pretraining on a few-shot problem of a higher way without fine-tuning does not necessarily improve the testing accuracy on the target problem of a lower way. As shown in Table~\ref{tab:cross-way-imagenet}, for the \emph{mini}ImageNet dataset, the testing accuracy of models trained with a higher way is similar or even worse than that trained with a lower way. In the original paper~\cite{snell2017prototypical}, the authors claimed that it is advantageous to use more classes (higher “way”) per training episode rather than fewer because a higher way always achieves higher accuracy than a lower way. Through our experiments, we found that this claim was biased because the authors only studied one learning rate decaying policy, which is too fast to converge to a near optimum for models trained with a lower way. When trained with a slow learning rate decaying policy, models trained with a lower way can catch up with or even surpass those trained with a higher way. 
	\item Pretraining with a higher way may generate a more ``universal'' feature representation, which helps to greatly speed up the convergence of the target problem with a lower-way and may improve the testing accuracy on some tasks. All considered models with pretraining can converge in less than 20000 iterations, which is much less than that of the target problems. In addition, for the \emph{mini}ImageNet $5$-way $1$-shot classification problem under Euclidean distance, pretraining on the $20$-way $1$-shot classification problem may achieve approximately $1\%$ accuracy improvement. However, we should also notice that the accuracies of the fine-tuned version of cross-way training are lower than those of non-fine-tuned versions on most tasks of Omniglot, which may due to the overfitting in fine-tuning the few-shot problems on this dataset. 
\end{itemize}
%Overall, training on a few-shot problem with a higher way converges faster than training on the target problem with a lower way because there are more data in each episode, which can speed up the convergence and improve the testing accuracy of the target problem when serving as a pretrainig technique.

\section{Conclusion and Future work}
This paper introduced a formal parallel between the regular classification problem and the few-shot classification problem from the perspective of supervised learning. 
% Quality control editor: Please ensure that the intended meaning has been maintained in the edits of the previous sentence.
On top of this formalism of parallel, we further propose multi-episode and cross-way training techniques, 
which correspond to the minibatch training and pretraining, respectively. 
The performance of multi-episode and cross-way training is guaranteed by the numerical experimental results on Prototypical Networks~\cite{snell2017prototypical} for varying few-shot classification problems of Omniglot and \emph{mini}ImageNet.
This research is in its early stage. There are several aspects that deserve deeper investigation: 
\begin{itemize}
	\item {\bf theoretically analyze} the performance on multi-episode and cross-way training of an artificial few-shot learning problem on a simple dataset;
	\item {\bf exploit} the performance on more few-shot classification approaches (such as Relation Networks~\cite{sung2018learning} and MAML~\cite{FinnAL2017Model}), as well as on more challenging dataset (such as tieredImageNet);
	\item {\bf design} the true minibatch and pretraining aspects for the few-shot classification problem, as that mentioned in Section~\ref{sec:methodology}; 
	\item {\bf explore} the influence of the proposed training approaches on the size of network architectures, because here, we view the size of training data as extremely large but not limited. 
\end{itemize}

\bibliographystyle{splncs04}
\bibliography{hmy}
\newpage

\begin{figure}[!htb]
	
	\centering
	\includegraphics[width=0.98\textwidth]{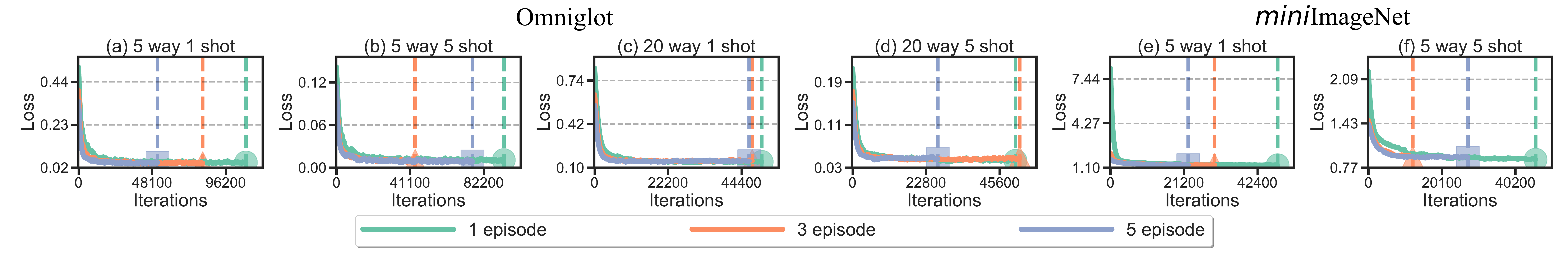}
	\caption{The validation loss curve of each multi-episode training case on Prototypical Networks~\cite{snell2017prototypical} (with Euclidean distance) with the optimum learning rate decaying policy for Omniglot and \emph{mini}ImageNet. The validation loss is averaged over 100 randomly generated episodes.} 
	\label{fig:multi-episode-loss-iter}
\end{figure}

\begin{figure}[!htb]
	\centering
	\iffalse
	\begin{tabular}{|m{0.05\textwidth}<{\centering}|m{0.05\textwidth}<{\centering}c|}
		\hline
		\multicolumn{1}{|c|}{\multirow{2}{*}{\rotatebox{90}{{\small Omniglot}}}} & {\rotatebox{90}{{\tiny Pre-train way=20}}&
			\mgape{\includegraphics[width=0.8\textwidth]{pic/crossway-x-logy-omniglot-20.png}}\\
			\cline{2-3}
			& \rotatebox{90}{{\tiny Pre-train way=60}}&
			\mgape{\includegraphics[width=0.8\textwidth]{pic/crossway-x-logy-omniglot-60.png}}\\
			\hline
			\multicolumn{1}{|c|}{\multirow{2}{*}{\rotatebox{90}{{\small \emph{mini}ImageNet}}}} & \rotatebox{90}{{\tiny Pre-train way=20}}&
			\mgape{\includegraphics[width=0.8\textwidth]{pic/crossway-x-logy-miniimagenet-20.png}}\\
			\cline{2-3}
			&\rotatebox{90}{{\tiny Pre-train way=30}}&
			\mgape{\includegraphics[width=0.8\textwidth]{pic/crossway-x-logy-miniimagenet-30.png}}\\
			\hline
		\end{tabular}
		\includegraphics[width=1\textwidth]{pic/Figure-cross.pdf}
		\fi
		\begin{tabular}{m{0.01\textwidth}<{\centering}cm{0.01\textwidth}<{\centering}c}
			&\multicolumn{1}{c}{\multirow{1}{*}{\scalebox{0.8}[0.8]{\rotatebox{0}{{\tiny \textbf{Omniglot}}}}}} & &
			\multicolumn{1}{c}{\multirow{1}{*}{\scalebox{0.8}[0.8]{\rotatebox{0}{{\tiny \textbf{\emph{mini}ImageNet}}}}}}\\
			\specialrule{0pt}{-2pt}{0pt}
			\scalebox{0.6}[0.6]{\rotatebox{90}{{\tiny \textbf{Pre-train way=20}}}}&
			\mgape{\includegraphics[width=0.47\textwidth]{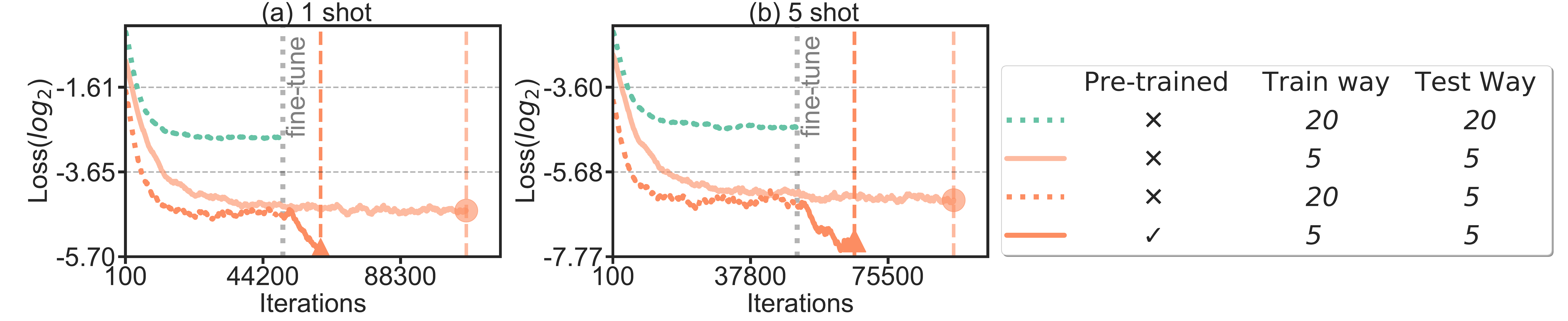}}
			&\scalebox{0.6}[0.6]{\rotatebox{90}{{\tiny \textbf{Pre-train way=20}}}}&
			\mgape{\includegraphics[width=0.47\textwidth]{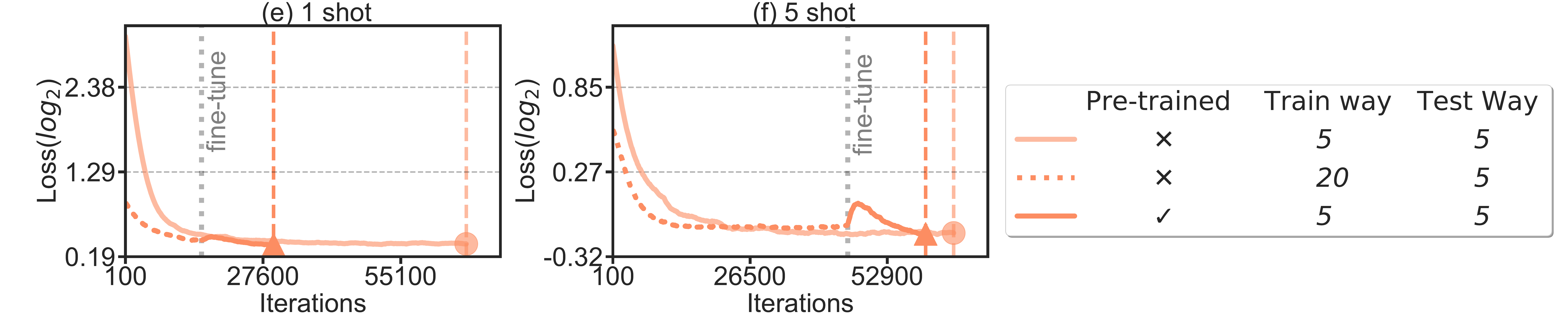}}\\
			\specialrule{0pt}{-2pt}{0pt}
			\scalebox{0.6}[0.6]{\rotatebox{90}{{\tiny \textbf{Pre-train way=60}}}}&
			\mgape{\includegraphics[width=0.47\textwidth]{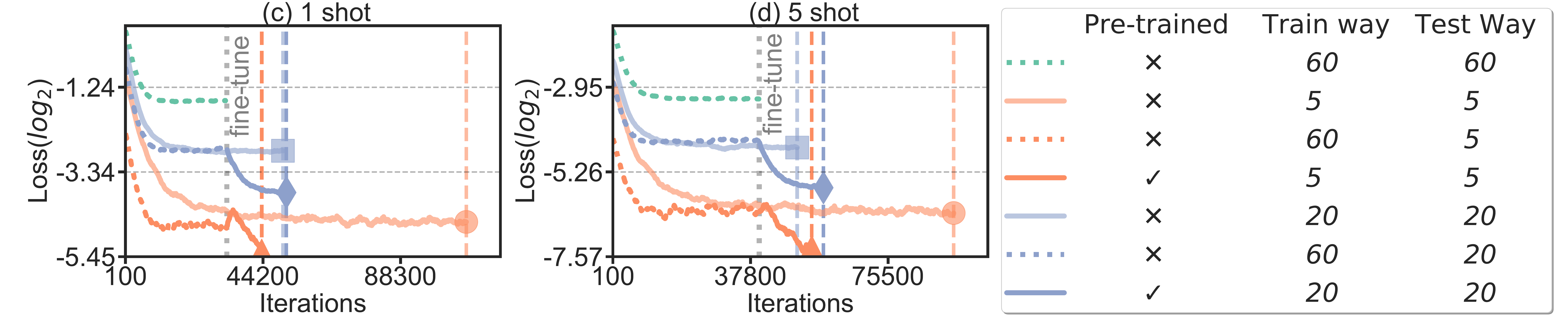}}
			& 
			\scalebox{0.6}[0.6]{\rotatebox{90}{{\tiny \textbf{Pre-train way=30}}}}&
			\mgape{\includegraphics[width=0.47\textwidth]{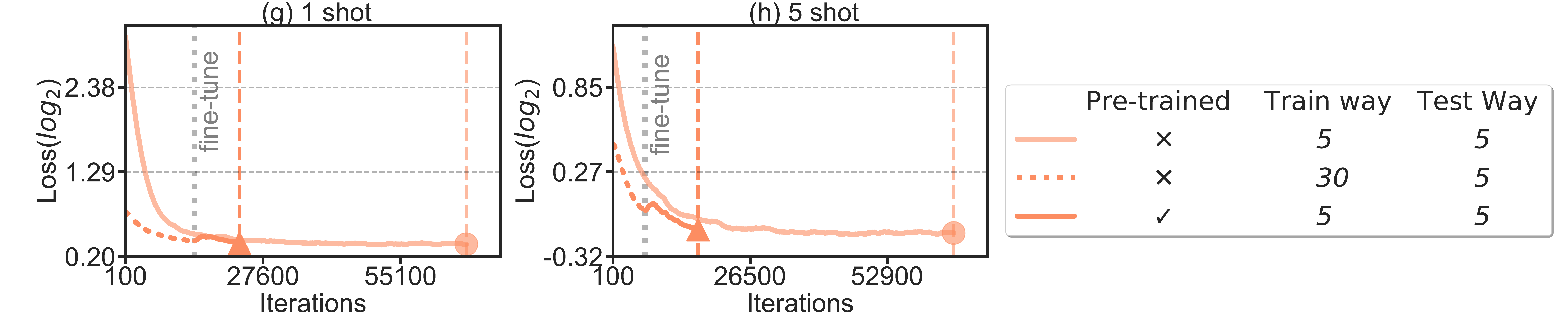}}\\
		\end{tabular}
		\caption{The validation loss curve of both the non-fine-tuned (N-FT) and fine-tuned (FT) versions of each cross-way training case on Prototypical Networks~\cite{snell2017prototypical} (with Euclidean distance) with the optimum learning rate decaying policy for Omniglot and \emph{mini}ImageNet. The validation loss is averaged over 100 randomly generated episodes. 
		}
		\label{fig:crossway-loss-iter}
		
	\end{figure}

\begin{figure}[!htb]
		\centering
		\includegraphics[width=0.4\textwidth]{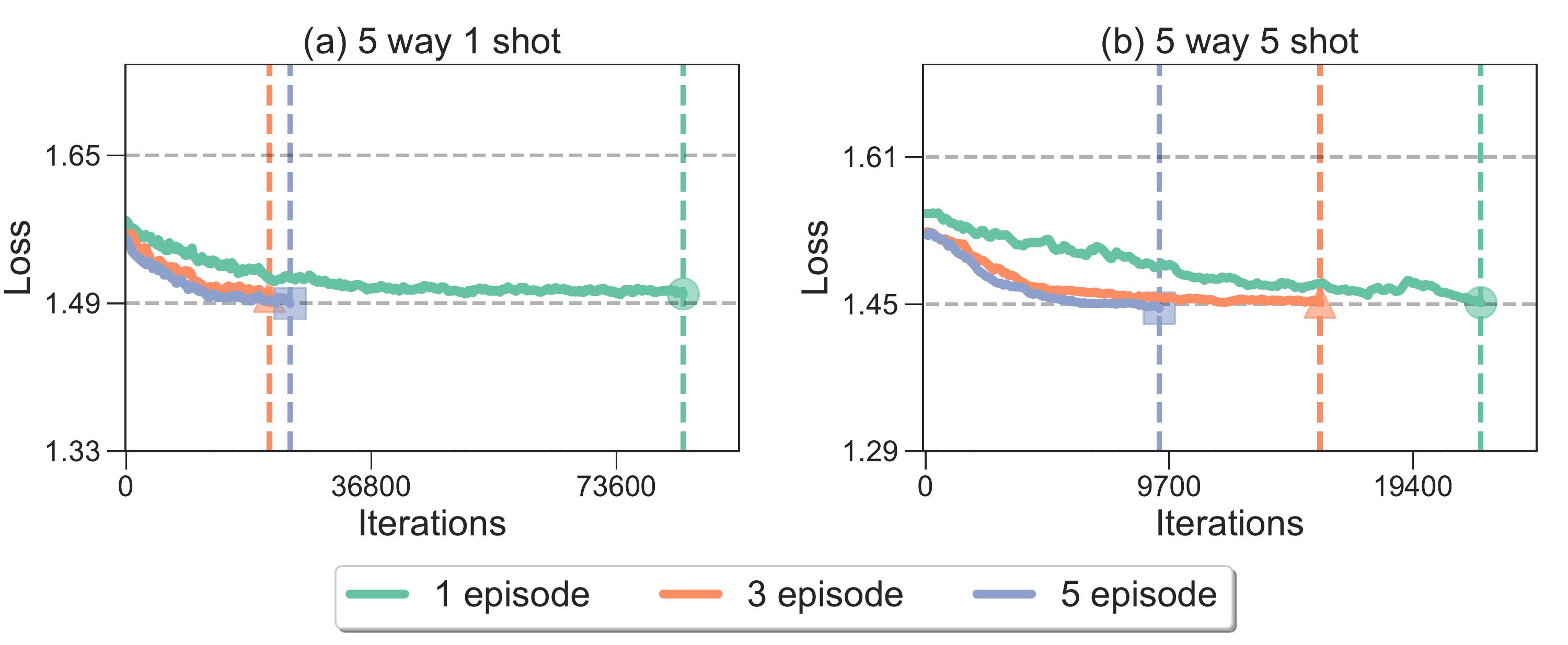}
		\caption{The validation loss curve of each multi-episode training case on Prototypical Networks~\cite{snell2017prototypical} (with cosine distance) with the optimum learning rate decaying policy for \emph{mini}ImageNet. The validation loss is averaged over 100 randomly generated episodes.} 
		\label{fig:multi-episode-loss-iter-cosine}
	\end{figure}

\begin{figure}[!htb]
		
		\centering
		\includegraphics[height=0.1\textheight]{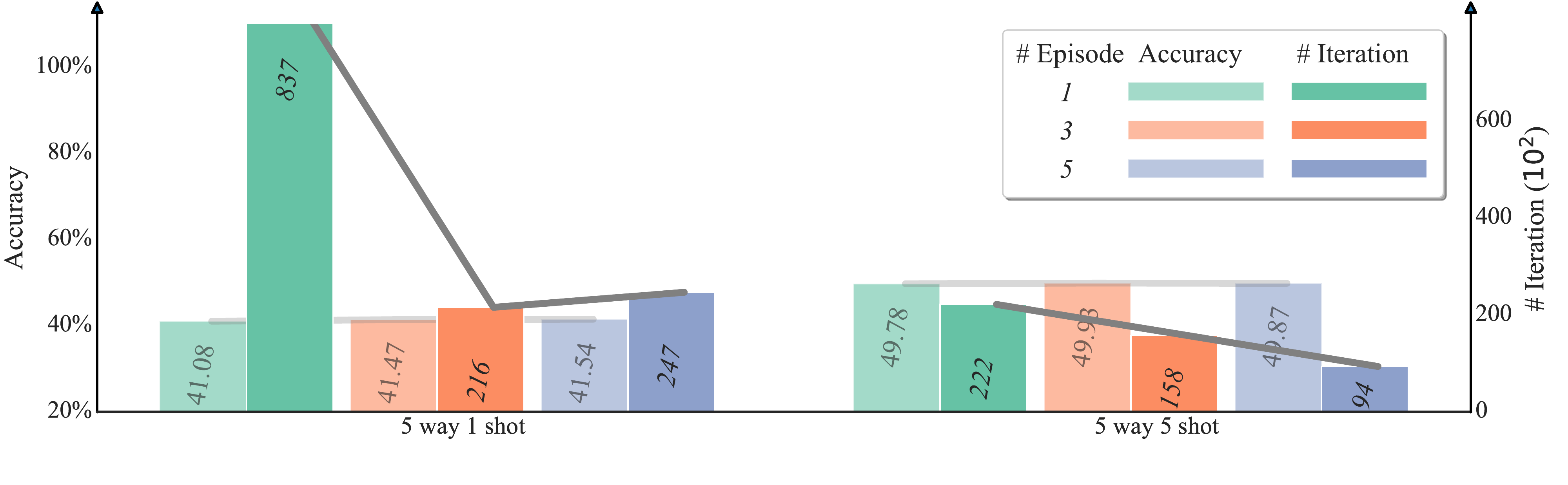} 
		\includegraphics[height=0.1\textheight]{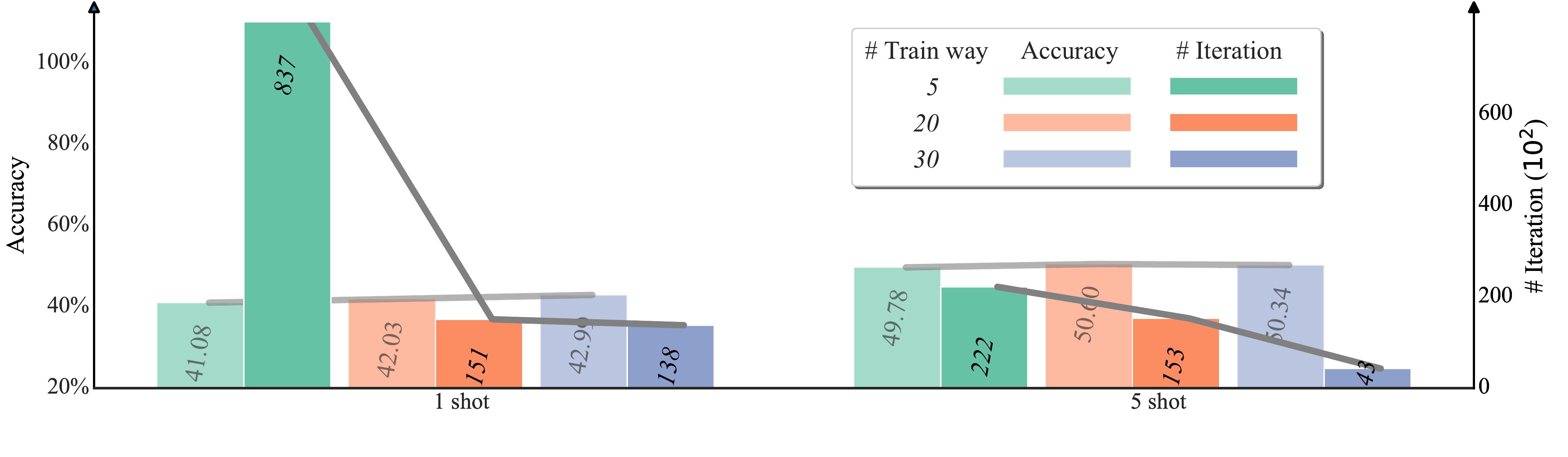} 
		\caption{The convergence accuracy-iterations results of \textbf{(Left)} multi-episode training and \textbf{(Right)} cross-way training on Prototypical Networks~\cite{snell2017prototypical} with Cosine distance with the optimum learning rate decaying policy for \emph{mini}ImageNet.}
		\label{fig:multi-episode-acc-iter-cosine}
	\end{figure}

\begin{figure}[!htb]
		\centering
		\begin{tabular}{m{0.01\textwidth}<{\centering}cm{0.01\textwidth}<{\centering}c}
			\scalebox{0.6}[0.6]{\rotatebox{90}{{\tiny \textbf{Pre-train way=20}}}}&
			\mgape{\includegraphics[width=0.47\textwidth]{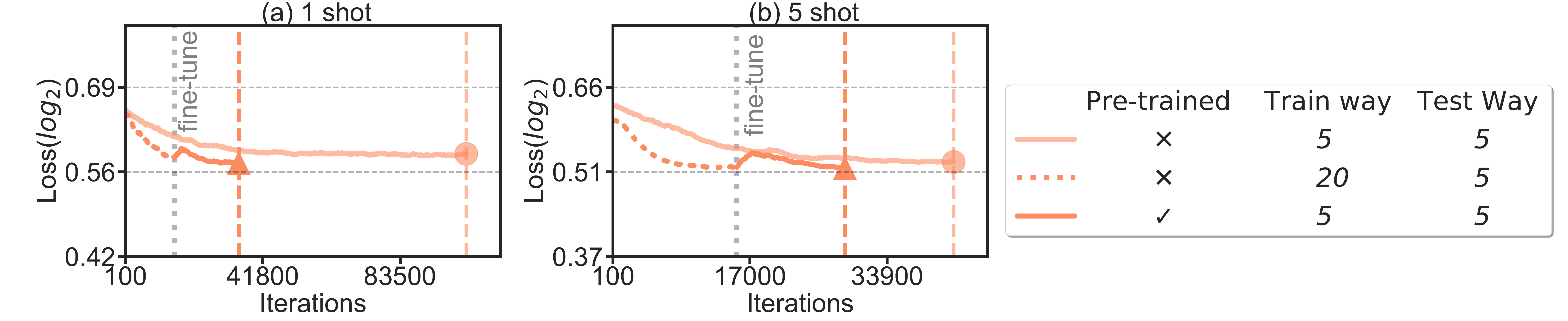}}
			& 
			\scalebox{0.6}[0.6]{\rotatebox{90}{{\tiny \textbf{Pre-train way=30}}}}&
			\mgape{\includegraphics[width=0.47\textwidth]{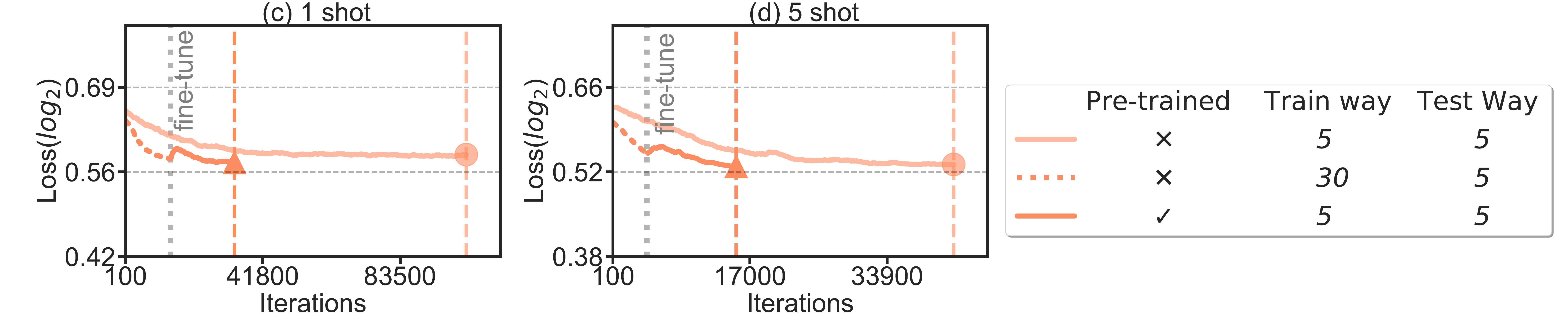}}\\
		\end{tabular}
		\caption{The validation loss curve of both the non-fine-tuned (N-FT) and fine-tuned (FT) versions of each cross-way training case on Prototypical Networks~\cite{snell2017prototypical} (with cosine distance) with the optimum learning rate decaying policy for \emph{mini}ImageNet. The validation loss is averaged over 100 randomly generated episodes. 
		}
		\label{fig:crossway-loss-iter-cosine}
	\end{figure}

\textbf{Supplementary Material.}
This supplementary material includes the detailed experimental results of the proposed multi-episode and cross-way training strategies on Prototypical Networks~\cite{snell2017prototypical} with Euclidean distance for Omniglot~\cite{Lake2011One} and the \emph{mini}ImageNet version of ILSVRC-2012~\cite{Russakovsky2015ImageNet} with the splits proposed by Ravi and Larochelle~\cite{ravi2017optimization}. The results of the two proposed training strategies on Prototypical Networks with cosine distance for \emph{mini}ImageNet are also included. The list of items are:
\begin{itemize}
	\item Table~\ref{tab:multi-episode} shows the convergence accuracy-iterations results of multi-episode training on Prototypical Networks~\cite{snell2017prototypical} (with Euclidean distance) with different episode numbers (extension of Table 2 and Table 3 of the main paper).
	\item Figure~\ref{fig:multi-episode-loss-iter} shows the validation loss curve of each multi-episode training case on Prototypical Networks~\cite{snell2017prototypical} (with Euclidean distance) with the optimum learning rate decaying policy.
	
	\item Table~\ref{tab:cross-way} shows the convergence accuracy-iterations results of the non-fine-tuned version of cross-way training on Prototypical Networks~\cite{snell2017prototypical} (with Euclidean distance)(extension of Table 4 and Table 5 of the main paper).
	\item Figure~\ref{fig:crossway-loss-iter} shows the validation loss curve of both the fine-tuned and non-fine-tuned versions of each cross-way training case on Prototypical Networks~\cite{snell2017prototypical} (with Euclidean distance) with the optimum learning rate decaying policy. 	
	\item Table~\ref{tab:multi-episode-cosine} shows the convergence accuracy-iterations results of multi-episode training on Prototypical Networks~\cite{snell2017prototypical} (with cosine distance) with different episode numbers (extension of Table 3 of the main paper).
	\item Figure~\ref{fig:multi-episode-loss-iter-cosine} shows the validation loss curve of each multi-episode training case on Prototypical Networks~\cite{snell2017prototypical} (with cosine distance) with the optimum learning rate decaying policy.
	\item Figure~\ref{fig:multi-episode-acc-iter-cosine} shows the convergence accuracy-iterations results of (\textbf{Left}) multi-episode training and (\textbf{Right}) cross-way training on Prototypical Networks~\cite{snell2017prototypical} (with cosine distance) with the optimum learning rate decaying policy.
	\item Table~\ref{tab:cross-way-cosine} shows the convergence accuracy-iterations results of the non-fine-tuned version of cross-way training on Prototypical Networks~\cite{snell2017prototypical} (with cosine distance) (extension of Table 5 of the main paper).
	\item Figure~\ref{fig:crossway-loss-iter-cosine} shows the validation loss curve of both the fine-tuned and non-fine-tuned versions of each cross-way training case on Prototypical Networks~\cite{snell2017prototypical} (with cosine distance) with the optimum learning rate decaying policy. 
\end{itemize}

\begin{table*}[!htb]
	\caption{The convergence accuracy-iterations results of multi-episode training on Prototypical Networks~\cite{snell2017prototypical} (with Euclidean distance) for Omniglot and \emph{mini}ImageNet. 
		The best-performing under different numbers of episodes is highlighted. Here, *results reported in the original paper~\cite{snell2017prototypical}. } \label{tab:multi-episode}
	\centering
	\begin{tiny}
		\scalebox{0.6}[0.6]{
			\begin{tabular}{l|c|c}
				\specialrule{1pt}{0pt}{2pt}
				\begin{tabular}{@{}l@{}}
					\specialrule{0pt}{0pt}{2.4pt}
					\multicolumn{1}{l}{\multirow{3}{*}{\textbf{Episodes}}} \\
					\\
					\\
					\specialrule{0.5pt}{4.4pt}{2pt}
					$\bm 1^*$ \\
					$\bm 1$\\
					$\bm 3$\\
					$\bm 5$\\
					\specialrule{0.8pt}{2pt}{4.4pt}

					\multicolumn{1}{l}{\multirow{3}{*}{\textbf{Episodes}}} \\
					\\
					\\
					\specialrule{0.5pt}{4.4pt}{2pt}
					$\bm 1^*$ \\
					$\bm 1$\\
					$\bm 3$\\
					$\bm 5$\\
					\specialrule{0.8pt}{2pt}{4.4pt}
					
					\multicolumn{1}{l}{\multirow{3}{*}{\textbf{Episodes}}} \\
					\\
					\\
					\specialrule{0.5pt}{4.4pt}{2pt}
					$\bm 1^*$ \\
					$\bm 1$\\
					$\bm 3$\\
					$\bm 5$\\		
				\end{tabular}&
				\begin{tabular}{@{}cccccc@{}}
					\multicolumn{6}{c}{\textbf{\textsc{Omniglot  5 Way 1 shot}}}\\
					\specialrule{0.8pt}{2pt}{2pt}
					\multicolumn{2}{c}{ $\bm 1 \times$ \textbf{Schedule}}          & \multicolumn{2}{c}{ $\bm 3 \times$ \textbf{Schedule}}   &  \multicolumn{2}{c}{ $\bm 5 \times$ \textbf{Schedule}} \\ 
					\textbf{Acc.} & \textbf{Iters.($\scriptstyle 10^2$)} & \textbf{Acc.} & \textbf{Iters.($\scriptstyle 10^2$)} & \textbf{Acc.} & \textbf{Iters.($\scriptstyle 10^2$)}    \\ 
					\specialrule{0.5pt}{2pt}{2pt}
					97.40\%          & -          & -  & -          & -  & -    
					\\ 
					97.19 $\pm$ 0.23\%&	353&	98.06 $\pm$ 0.19\%&	618&	\textbf{98.28 $\pm$ 0.19\%}&	\textbf{1094}\\
					97.96 $\pm$ 0.19\%&	415&	98.45 $\pm$ 0.17\%&	507&	\textbf{98.49 $\pm$ 0.17\%}&	\textbf{812}\\
					98.21 $\pm$ 0.19\%&	517&	\textbf{98.56 $\pm$ 0.16\%}&	\textbf{517}&	{98.50 $\pm$ 0.17\%}&	{517}\\		
					\specialrule{0.8pt}{2pt}{2pt}		
					\multicolumn{6}{c}{\textbf{\textsc{Omniglot  20 Way 1 shot}}}\\
					\specialrule{0.8pt}{2pt}{2pt}
					\multicolumn{2}{c}{ $\bm 1 \times$ \textbf{Schedule} }          & \multicolumn{2}{c}{ $\bm 3 \times$ \textbf{Schedule}}   &  \multicolumn{2}{c}{ $\bm 5 \times$ \textbf{Schedule}} \\ 
					\textbf{Acc.} & \textbf{Iters.($\scriptstyle 10^2$)} & \textbf{Acc.} & \textbf{Iters.($\scriptstyle 10^2$)} & \textbf{Acc.} & \textbf{Iters.($\scriptstyle 10^2$)}    \\ 
					\specialrule{0.5pt}{2pt}{2pt}
					95.40\%          & -          & -  & -          & -  & - \\     
					
					94.37 $\pm$ 0.18\%&	506	&94.94 $\pm$ 0.17\%&	506&	\textbf{95.10 $\pm$ 0.17\%}&	\textbf{506}	\\
					94.82 $\pm$ 0.17\%&	579&	\textbf{95.11 $\pm$ 0.17\%}&	\textbf{477}&	95.01 $\pm$ 0.17\%&	422\\
					94.68 $\pm$ 0.18\%&	319&	\textbf{94.98 $\pm$ 0.17\%}&	\textbf{468}&	94.94 $\pm$ 0.17\%&	325\\
					\specialrule{0.8pt}{2pt}{2pt}
					\multicolumn{6}{c}{\textbf{\textsc{\emph{mini}ImageNet  5 Way 1 shot}}}\\
					\specialrule{0.8pt}{2pt}{2pt}
					\multicolumn{2}{c}{ $\bm 1 \times$ \textbf{Schedule} }          & \multicolumn{2}{c}{ $\bm 3 \times$ \textbf{Schedule}}   &  \multicolumn{2}{c}{ $\bm 5 \times$ \textbf{Schedule}} \\ 
					\textbf{Acc.} & \textbf{Iters.($\scriptstyle 10^2$)} & \textbf{Acc.} & \textbf{Iters.($\scriptstyle 10^2$)} & \textbf{Acc.} & \textbf{Iters.($\scriptstyle 10^2$)}    \\ 
					\specialrule{0.5pt}{2pt}{2pt}		
					46.61 $\pm$ 0.78\%          & -          & -  & -          & -  & -      
					\\ 	
					43.04 $\pm$ 0.74\%&	282&	47.53 $\pm$ 0.81\%	&537&	\textbf{48.39 $\pm$ 0.80\%}&	\textbf{483}\\
					46.30 $\pm$ 0.79\%&	301&	48.54 $\pm$ 0.80\%&	223&	\textbf{48.81 $\pm$ 0.79\%}&	\textbf{301}\\
					47.42 $\pm$ 0.78\%&	265&	48.52 $\pm$ 0.77\%&	225&	\textbf{49.00 $\pm$ 0.80\%}&	\textbf{225}\\   
				\end{tabular}&
				\begin{tabular}{@{}cccccc@{}}
					\multicolumn{6}{c}{\textbf{\textsc{Omniglot  5 Way 5 shot}}}\\
					\specialrule{0.8pt}{2pt}{2pt}
					\multicolumn{2}{c}{ $\bm 1 \times$ \textbf{Schedule} }          & \multicolumn{2}{c}{ $\bm 3 \times$ \textbf{Schedule}}   &  \multicolumn{2}{c}{ $\bm 5 \times$ \textbf{Schedule}} \\ 
					\textbf{Acc.} & \textbf{Iters.($\scriptstyle 10^2$)} & \textbf{Acc.} & \textbf{Iters.($\scriptstyle 10^2$)} & \textbf{Acc.} & \textbf{Iters.($\scriptstyle 10^2$)}    \\ 
					\specialrule{0.5pt}{2pt}{2pt}    	
					99.30\%  & -&- & -&- &-  \\ 	
					99.31 $\pm$ 0.09\%&	935&	99.49 $\pm$ 0.08\%&	730&	\textbf{99.53 $\pm$ 0.07\%}&	\textbf{935}\\
					99.51 $\pm$ 0.08\%&	310&	\textbf{99.59 $\pm$ 0.07\%}&	\textbf{440}&	\textbf{99.59 $\pm$ 0.07\%}&	\textbf{440}\\
					99.51 $\pm$ 0.08\%&	370&	99.59 $\pm$ 0.07\%&	357&	\textbf{99.61 $\pm$ 0.07\%}&	\textbf{760}\\
					\specialrule{0.8pt}{2pt}{2pt}
					\multicolumn{6}{c}{\textbf{\textsc{Omniglot  20 Way 5 shot}}}\\
					\specialrule{0.8pt}{2pt}{2pt}
					\multicolumn{2}{c}{ $\bm 1 \times$ \textbf{Schedule} }          & \multicolumn{2}{c}{ $\bm 3 \times$ \textbf{Schedule}}   &  \multicolumn{2}{c}{ $\bm 5 \times$ \textbf{Schedule}} \\ 
					\textbf{Acc.} & \textbf{Iters.($\scriptstyle 10^2$)} & \textbf{Acc.} & \textbf{Iters.($\scriptstyle 10^2$)} & \textbf{Acc.} & \textbf{Iters.($\scriptstyle 10^2$)}    \\ 
					\specialrule{0.5pt}{2pt}{2pt}
					98.70\%  & -&- & -&- &- \\  	
					98.51 $\pm$ 0.06\%&	506&	98.68 $\pm$ 0.06\%&	702&	\textbf{98.70 $\pm$ 0.06\%}&	\textbf{506}\\
					98.62 $\pm$ 0.06\%&	519&	\textbf{98.75 $\pm$ 0.06\%}&	\textbf{519}&	98.71 $\pm$ 0.06\%&	448\\
					98.66 $\pm$ 0.06\%&	442&	\textbf{98.70 $\pm$ 0.06\%}&	\textbf{265}&	98.64 $\pm$ 0.06\%&	317\\
					\specialrule{0.8pt}{2pt}{2pt}
					\multicolumn{6}{c}{\textbf{\textsc{\emph{mini}ImageNet  5 Way 5 shot}}}\\
					\specialrule{0.8pt}{2pt}{2pt}
					\multicolumn{2}{c}{ $\bm 1 \times$ \textbf{Schedule} }          & \multicolumn{2}{c}{ $\bm 3 \times$ \textbf{Schedule}}   &  \multicolumn{2}{c}{ $\bm 5 \times$ \textbf{Schedule}} \\ 
					\textbf{Acc.} & \textbf{Iters.($\scriptstyle 10^2$)} & \textbf{Acc.} & \textbf{Iters.($\scriptstyle 10^2$)} & \textbf{Acc.} & \textbf{Iters.($\scriptstyle 10^2$)}    \\ 
					\specialrule{0.5pt}{2pt}{2pt}
					65.77 $\pm$ 0.70\%  & -&- & -&- &- \\
					64.03 $\pm$ 0.69\%&	125	&66.10 $\pm$ 0.64\%&	439&	\textbf{66.24 $\pm$ 0.65\%}&	\textbf{458}\\
					65.03 $\pm$ 0.68\%&	276	&\textbf{65.85 $\pm$ 0.65\%}&	\textbf{122}	&65.41 $\pm$ 0.65\%	&143\\
					\textbf{65.77 $\pm$ 0.67\%}&	\textbf{273}&	65.09 $\pm$ 0.68\%&	96&	65.36 $\pm$ 0.68\%&	96\\   
				\end{tabular}\\
				\specialrule{1pt}{2pt}{0pt}
			\end{tabular}
		}
	\end{tiny}
\end{table*}

\begin{table*}[!htb]
	\caption{The convergence accuracy-iterations results of the non-fine-tuned version of cross-way training on Prototypical Networks~\cite{snell2017prototypical} (with Euclidean distance) for Omniglot and \emph{mini}ImageNet.
		Here, *results reported in the original paper~\cite{snell2017prototypical}. } \label{tab:cross-way}
	\begin{center}
		\begin{tiny}
			\scalebox{0.7}[0.7]{	
				\begin{tabular}{cccccccccccc}	
					\specialrule{1pt}{0pt}{2pt}
					\multicolumn{12}{c}{\textbf{\textsc{Omniglot 1 shot}}}                        \\ 
					\specialrule{0.8pt}{2pt}{2pt}
					\multicolumn{1}{c}{\multirow{3}{*}{$\hat K$}}	& \multicolumn{2}{c}{ $\bm 1 \times$ \textbf{Schedule} } &\multicolumn{3}{c}{$\bm 1 \times$ \textbf{Schedule} }          & \multicolumn{3}{c}{$\bm 3 \times$ \textbf{Schedule}}   &  \multicolumn{3}{c}{$\bm 5 \times$ \textbf{Schedule}}  	 \\
					& \multicolumn{2}{c}{\textbf{Acc.*}} 	& \multicolumn{2}{c}{\textbf{Acc.}} & \multicolumn{1}{c}{\multirow{1}{*}{\textbf{Iters.($\scriptstyle 10^2$)}}}& \multicolumn{2}{c}{\textbf{Acc.}} & \multicolumn{1}{c}{\multirow{1}{*}{\textbf{Iters.($\scriptstyle 10^2$)}}} &\multicolumn{2}{c}{\textbf{Acc.}} & \multicolumn{1}{c}{\multirow{1}{*}{\textbf{Iters.($\scriptstyle 10^2$)}}}           \\ 
					
					&$K=5$ & $K=20$ 	&$K=5$ & $K=20$ & &$K=5$ & $K=20$ & &$K=5$ & $K=20$ & \\
					\specialrule{0.5pt}{2pt}{2pt}
					$\bm 5$  &97.4\% & 92.0\%
					&97.19 $\pm$ 0.23\%&	90.83 $\pm$ 0.23\%&	353&	98.06 $\pm$ 0.19\%&	93.28 $\pm$ 0.20\%&	618&	\textbf{98.28 $\pm$ 0.19\%}&	\textbf{93.89 $\pm$ 0.19\%}&	\textbf{1094}\\
					$\bm 20$ &98.7\% &95.4\% &98.46 $\pm$ 0.17\%&	94.37 $\pm$ 0.18\%&	506&	98.67 $\pm$ 0.16\%&	94.94 $\pm$ 0.17\%&	506&	\textbf{98.70 $\pm$ 0.16\%}&	\textbf{95.10 $\pm$ 0.17\%}&	\textbf{506}\\
					$\bm 60$ &98.8\% &96.0\%
					&98.66 $\pm$ 0.15\%&	95.16 $\pm$ 0.17\%&	462&	\textbf{98.73 $\pm$ 0.15\%}&	\textbf{95.21 $\pm$ 0.17\%}&	\textbf{324}&	98.71 $\pm$ 0.16\%&	95.11 $\pm$ 0.17\%&	324
					\\
					\specialrule{0.8pt}{2pt}{2pt}
					\multicolumn{12}{c}{\textbf{\textsc{Omniglot 5 shot}}}                         \\ 
					\specialrule{0.8pt}{2pt}{2pt}
					\multicolumn{1}{c}{\multirow{3}{*}{$\hat K$}}	& \multicolumn{2}{c}{ $\bm 1 \times$ \textbf{Schedule} } &\multicolumn{3}{c}{$\bm 1 \times$ \textbf{Schedule} }          & \multicolumn{3}{c}{$\bm 3 \times$ \textbf{Schedule}}   &  \multicolumn{3}{c}{$\bm 5 \times$ \textbf{Schedule}}  	 \\
					& \multicolumn{2}{c}{\textbf{Acc.*}} 	& \multicolumn{2}{c}{\textbf{Acc.}} & \multicolumn{1}{c}{\multirow{1}{*}{\textbf{Iters.($\scriptstyle 10^2$)}}}& \multicolumn{2}{c}{\textbf{Acc.}} & \multicolumn{1}{c}{\multirow{1}{*}{\textbf{Iters.($\scriptstyle 10^2$)}}} &\multicolumn{2}{c}{\textbf{Acc.}} & \multicolumn{1}{c}{\multirow{1}{*}{\textbf{Iters.($\scriptstyle 10^2$)}}}           \\ 
					
					&$K=5$ & $K=20$ 	&$K=5$ & $K=20$ & &$K=5$ & $K=20$ & &$K=5$ & $K=20$ & \\
					\specialrule{0.5pt}{2pt}{2pt}
					
					$\bm 5$ &99.3\% & 97.8\% &99.31 $\pm$ 0.09\%&	97.45 $\pm$ 0.08\%&	937&	99.49 $\pm$ 0.08\%&	98.20 $\pm$ 0.07\%&	730&	\textbf{99.53 $\pm$ 0.07\%}&	\textbf{98.36 $\pm$ 0.06\%}&	\textbf{935}\\
					$\bm 20$ &99.6\% & 98.7\% &99.59 $\pm$ 0.08\%&	98.51 $\pm$ 0.06\%&	506&	99.62 $\pm$ 0.07\%&	98.68 $\pm$ 0.06\%&	702&	\textbf{99.63 $\pm$ 0.07\%}&	\textbf{98.70 $\pm$ 0.06\%}&	\textbf{506}\\
					$\bm 60$ &99.7\% & 98.9\%&99.65 $\pm$ 0.07\%&	98.66 $\pm$ 0.06\%&	463&	\textbf{99.64 $\pm$ 0.07\%}&	\textbf{98.77 $\pm$ 0.05\%}&	\textbf{402}&	99.62 $\pm$ 0.07\%&	98.70 $\pm$ 0.06\%&	310\\
					\specialrule{0.8pt}{2pt}{2pt}
					\multicolumn{12}{c}{\textbf{\textsc{\emph{mini}ImageNet 1 shot}}}                         \\ 
					\specialrule{0.8pt}{2pt}{2pt}
					\multicolumn{1}{c}{\multirow{3}{*}{$\hat K$}}& \multicolumn{2}{c}{ $\bm 1 \times$ \textbf{Schedule} } &\multicolumn{3}{c}{$\bm 1 \times$ \textbf{Schedule} }          & \multicolumn{3}{c}{$\bm 3 \times$ \textbf{Schedule}}   &  \multicolumn{3}{c}{$\bm 5 \times$ \textbf{Schedule}} \\
					& \multicolumn{2}{c}{\textbf{Acc.*}} 	& \multicolumn{2}{c}{\textbf{Acc.}} & \multicolumn{1}{c}{\multirow{1}{*}{\textbf{Iters.($\scriptstyle 10^2$)}}}& \multicolumn{2}{c}{\textbf{Acc.}} & \multicolumn{1}{c}{\multirow{1}{*}{\textbf{Iters.($\scriptstyle 10^2$)}}} &\multicolumn{2}{c}{\textbf{Acc.}} & \multicolumn{1}{c}{\multirow{1}{*}{\textbf{Iters.($\scriptstyle 10^2$)}}}           \\ 
					
					& \multicolumn{2}{c}{$K=5$}	& \multicolumn{2}{c}{$K=5$}  & & \multicolumn{2}{c}{$K=5$}  & &\multicolumn{2}{c}{$K=5$}  & \\
					\specialrule{0.5pt}{2pt}{2pt} 
					
					$\bm 5$ & \multicolumn{2}{c}{46.61 $\pm$ 0.78\%}& \multicolumn{2}{c}{43.04 $\pm$ 0.74\%}&	282&	\multicolumn{2}{c}{47.53 $\pm$ 0.81\%}&	537&	\multicolumn{2}{c}{\textbf{48.39 $\pm$ 0.79\%}}&	\textbf{483}\\
					
					$\bm 20$ & \multicolumn{2}{c}{48.57 $\pm$ 0.79\%}&\multicolumn{2}{c}{46.93 $\pm$ 0.79\%}&	153&	\multicolumn{2}{c}{48.14 $\pm$ 0.78\%}&	153&	\multicolumn{2}{c}{\textbf{48.20 $\pm$ 0.80\%}}&	\textbf{153}\\
					
					$\bm 30$ & \multicolumn{2}{c}{49.42 $\pm$ 0.78\%} &\multicolumn{2}{c}{47.03$\pm$ 0.82\%}&	105&	\multicolumn{2}{c}{\textbf{48.37$\pm$ 0.77\%}}&	\textbf{138}&	\multicolumn{2}{c}{48.16 $\pm$ 0.78\%}&	160\\
					\specialrule{0.8pt}{2pt}{2pt}
					\multicolumn{12}{c}{\textbf{\textsc{\emph{mini}ImageNet 5 shot}}}                         \\ 
					\specialrule{0.8pt}{2pt}{2pt}
					\multicolumn{1}{c}{\multirow{3}{*}{$\hat K$}}& \multicolumn{2}{c}{ $\bm 1 \times$ \textbf{Schedule} } &\multicolumn{3}{c}{$\bm 1 \times$ \textbf{Schedule} }          & \multicolumn{3}{c}{$\bm 3 \times$ \textbf{Schedule}}   &  \multicolumn{3}{c}{$\bm 5 \times$ \textbf{Schedule}} \\
					& \multicolumn{2}{c}{\textbf{Acc.*}} 	& \multicolumn{2}{c}{\textbf{Acc.}} & \multicolumn{1}{c}{\multirow{1}{*}{\textbf{Iters.($\scriptstyle 10^2$)}}}& \multicolumn{2}{c}{\textbf{Acc.}} & \multicolumn{1}{c}{\multirow{1}{*}{\textbf{Iters.($\scriptstyle 10^2$)}}} &\multicolumn{2}{c}{\textbf{Acc.}} & \multicolumn{1}{c}{\multirow{1}{*}{\textbf{Iters.($\scriptstyle 10^2$)}}}           \\ 
					
					& \multicolumn{2}{c}{$K=5$}	& \multicolumn{2}{c}{$K=5$}  & & \multicolumn{2}{c}{$K=5$}  & &\multicolumn{2}{c}{$K=5$}  & \\
					\specialrule{0.5pt}{2pt}{2pt} 
					$\bm 5$& \multicolumn{2}{c}{65.77 $\pm$ 0.70\%}&	\multicolumn{2}{c}{64.03 $\pm$ 0.69\%}&	125&	\multicolumn{2}{c}{66.10 $\pm$ 0.64\%}&	439&\multicolumn{2}{c}{\textbf{66.24 $\pm$ 0.65\%}}&	\textbf{458}\\
					$\bm 20$	& \multicolumn{2}{c}{68.20 $\pm$ 0.66\%}&	\multicolumn{2}{c}{\textbf{65.68 $\pm$ 0.66\%}}&	\textbf{453}&	\multicolumn{2}{c}{64.96 $\pm$ 0.68\%}&	80&\multicolumn{2}{c}{63.50 $\pm$ 0.67\%}&	153\\
					
					$\bm 30$	& \multicolumn{2}{c}{66.79 $\pm$ 0.66\%} &	\multicolumn{2}{c}{64.56 $\pm$ 0.67\%}&	117&	\multicolumn{2}{c}{\textbf{65.24 $\pm$ 0.69\%}}&	\textbf{63}&	\multicolumn{2}{c}{61.70 $\pm$ 0.65\%}&	117\\
					\specialrule{1pt}{0pt}{2pt} 
					
			\end{tabular} }
		\end{tiny}
	\end{center}
\end{table*}
\setlength{\tabcolsep}{1pt}

	\begin{table*}[!htb]
		\caption{The convergence accuracy-iterations results of multi-episode training on Prototypical Networks~\cite{snell2017prototypical} (with cosine distance) for \emph{mini}ImageNet. 
			The best-performing under different numbers of episodes is highlighted. Here, *results reported in the original paper~\cite{snell2017prototypical}. } \label{tab:multi-episode-cosine}
		\centering
		\begin{tiny}
			\scalebox{0.7}[0.7]{
				\begin{tabular}{l|c|c}
					\specialrule{1pt}{0pt}{2pt}
					\begin{tabular}{@{}l@{}}
						\specialrule{0pt}{0pt}{2.4pt}
						\multicolumn{1}{l}{\multirow{3}{*}{\textbf{Episodes}}} \\
						\\
						\\
						\specialrule{0.5pt}{4.4pt}{2pt}
						$\bm 1^*$ \\
						$\bm 1$\\
						$\bm 3$\\
						$\bm 5$\\		
					\end{tabular}&
					\begin{tabular}{@{}cccccc@{}}
						
						\multicolumn{6}{c}{\textbf{\textsc{							5 Way 1 shot}}}\\
						\specialrule{0.8pt}{2pt}{2pt}
						\multicolumn{2}{c}{ $\bm 1 \times$ \textbf{Schedule} }          & \multicolumn{2}{c}{ $\bm 3 \times$ \textbf{Schedule}}   &  \multicolumn{2}{c}{ $\bm 5 \times$ \textbf{Schedule}} \\ 
						\textbf{Acc.} & \textbf{Iters.($\scriptstyle 10^2$)} & \textbf{Acc.} & \textbf{Iters.($\scriptstyle 10^2$)} & \textbf{Acc.} & \textbf{Iters.($\scriptstyle 10^2$)}    \\ 
						\specialrule{0.5pt}{2pt}{2pt}		
						38.82 $\pm$ 0.69\%
						& -          & -  & -          & -  & -      
						\\ 	
						39.30 $\pm$ 0.73\%&	282&	39.49 $\pm$ 0.72\%&	169	&\textbf{41.08 $\pm$ 0.72\%}&	\textbf{837}\\
						40.64 $\pm$ 0.73\%&	216	&40.92 $\pm$ 0.72\%&	216	& \textbf{41.47 $\pm$ 0.71\%}	&\textbf{216}\\
						40.89 $\pm$ 0.72\%&	453&	41.02 $\pm$ 0.68\%&	453	&\textbf{41.54 $\pm$ 0.73\%}&	\textbf{247}\\
						
					\end{tabular}&
					\begin{tabular}{@{}cccccc@{}}
						\multicolumn{6}{c}{\textbf{\textsc{5 Way 5 shot}}}\\
						\specialrule{0.8pt}{2pt}{2pt}
						\multicolumn{2}{c}{ $\bm 1 \times$ \textbf{Schedule} }          & \multicolumn{2}{c}{ $\bm 3 \times$ \textbf{Schedule}}   &  \multicolumn{2}{c}{ $\bm 5 \times$ \textbf{Schedule}} \\ 
						\textbf{Acc.} & \textbf{Iters.($\scriptstyle 10^2$)} & \textbf{Acc.} & \textbf{Iters.($\scriptstyle 10^2$)} & \textbf{Acc.} & \textbf{Iters.($\scriptstyle 10^2$)}    \\ 
						\specialrule{0.5pt}{2pt}{2pt}
						51.23 $\pm$0.63\%
						& -&- & -&- &- \\
						47.90 $\pm$ 0.64\%&	125&	49.22 $\pm$ 0.67\%&	220&	\textbf{49.78 $\pm$ 0.67\%}&	\textbf{222}\\
						\textbf{49.93 $\pm$ 0.68\%}&	\textbf{158}&	49.55 $\pm$ 0.70\%&	143&	48.68 $\pm$ 0.67\%&	143\\
						\textbf{49.87 $\pm$ 0.70\%}&	\textbf{94}&	49.26 $\pm$ 0.70\%&	94&	48.50 $\pm$ 0.67\%&	94\\
						
					\end{tabular}\\
					\specialrule{1pt}{2pt}{0pt}
				\end{tabular}
			}
		\end{tiny}
	\end{table*}
	
	\vspace{3cm}
	\begin{table*}[!htb]
		\caption{The convergence accuracy-iterations results of the non-fine-tuned version of cross-way training on Prototypical Networks~\cite{snell2017prototypical} (with cosine distance) for \emph{mini}ImageNet.
			Here, *results reported in the original paper~\cite{snell2017prototypical}. } \label{tab:cross-way-cosine}
		\begin{center}
			\begin{tiny}
				\scalebox{0.55}[0.55]{	
					\begin{tabular}{c|c|c}
						\specialrule{1pt}{0pt}{2pt}
						\begin{tabular}{@{}c@{}}
							\specialrule{0pt}{0pt}{2.4pt}
							\multicolumn{1}{c}{\multirow{3}{*}{\textbf{Train Way}}} \\
							\\
							\\
							\specialrule{0.5pt}{4.4pt}{2pt}
							$\bm 5$\\
							$\bm 20$\\
							$\bm 30$\\		
						\end{tabular}&
						\begin{tabular}{@{}ccccccc@{}}
							
							\multicolumn{7}{c}{\textbf{\textsc{							5 Way 1 shot}}}\\
							\specialrule{0.8pt}{2pt}{2pt}
							\multicolumn{1}{c}{ $\bm 1 \times$ \textbf{Schedule} }&\multicolumn{2}{c}{ $\bm 1 \times$ \textbf{Schedule} }          & \multicolumn{2}{c}{ $\bm 3 \times$ \textbf{Schedule}}   &  \multicolumn{2}{c}{ $\bm 5 \times$ \textbf{Schedule}} \\ 
							\textbf{Acc.*} &\textbf{Acc.} & \textbf{Iters.($\scriptstyle 10^2$)} & \textbf{Acc.} & \textbf{Iters.($\scriptstyle 10^2$)} & \textbf{Acc.} & \textbf{Iters.($\scriptstyle 10^2$)}    \\ 
							\specialrule{0.5pt}{2pt}{2pt}		
							38.82  $\pm$ 0.69\%& 	39.30 $\pm$ 0.73\%& 	282& 	39.49 $\pm$ 0.72\%& 	169& 	\textbf{41.08 $\pm$ 0.72\%}& 	\textbf{837}\\
							43.63 $\pm$ 0.76\%& 	41.73 $\pm$ 0.75\%& 	153& 	41.97 $\pm$ 0.79\%& 	151&	\textbf{42.03 $\pm$ 0.76\%}&	\textbf{151}\\
							-&42.08 $\pm$ 0.75\%&	138&	\textbf{42.99 $\pm$ 0.78\%}&	\textbf{138}&	42.38 $\pm$ 0.77\%&	138\\

						\end{tabular}&
						\begin{tabular}{@{}ccccccc@{}}
							\multicolumn{6}{c}{\textbf{\textsc{5 Way 5 shot}}}\\
							\specialrule{0.8pt}{2pt}{2pt}
							\multicolumn{1}{c}{ $\bm 1 \times$ \textbf{Schedule} }&\multicolumn{2}{c}{ $\bm 1 \times$ \textbf{Schedule} }          & \multicolumn{2}{c}{ $\bm 3 \times$ \textbf{Schedule}}   &  \multicolumn{2}{c}{ $\bm 5 \times$ \textbf{Schedule}} \\ 
							\textbf{Acc.*} &\textbf{Acc.} & \textbf{Iters.($\scriptstyle 10^2$)} & \textbf{Acc.} & \textbf{Iters.($\scriptstyle 10^2$)} & \textbf{Acc.} & \textbf{Iters.($\scriptstyle 10^2$)}    \\ 
							\specialrule{0.5pt}{2pt}{2pt}
							51.23 $\pm$ 0.63\%&	47.90 $\pm$ 0.64\%&	125&	49.22 $\pm$ 0.67\%&	220&	\textbf{49.78 $\pm$ 0.67\%}&	\textbf{222}\\
							51.48 $\pm$ 0.70\%&	\textbf{50.60 $\pm$ 0.68\%}&	\textbf{153}&	48.49 $\pm$ 0.72\%&	153&	48.07 $\pm$ 0.70\%&	153\\
							-	&\textbf{50.34 $\pm$ 0.68\%}&	\textbf{43}&	49.50 $\pm$ 0.69\%&	63&	49.44 $\pm$ 0.67\%&	43\\

						\end{tabular}\\
						\specialrule{1pt}{2pt}{0pt}
					\end{tabular}
				}
			\end{tiny}
		\end{center}
	\end{table*}
	
	\setlength{\tabcolsep}{1pt}
%\bibliographystyle{plain}
%\bibliography{example}

\end{document}